\documentclass{article}

\usepackage{tabularx} %

\usepackage{microtype}
\usepackage{graphicx}
\usepackage{subfigure}
\usepackage{booktabs} %
\usepackage[dvipsnames]{xcolor}
\usepackage{mdframed} %
\usepackage{amssymb}%
\usepackage{pifont}%
\usepackage{listings}
\usepackage{multirow} 
\usepackage{caption}
\usepackage{subcaption}
\usepackage{colortbl}
\usepackage{enumitem}
\definecolor{codeblue}{rgb}{0.25,0.5,0.5}
\definecolor{codekw}{rgb}{0.85, 0.18, 0.50}
\definecolor{codegreen}{rgb}{0,0.6,0}
\definecolor{codegray}{rgb}{0.5,0.5,0.5}
\definecolor{codepurple}{rgb}{0.58,0,0.82}
\usepackage{wrapfig,lipsum}
\usepackage{threeparttable}
\usepackage{ulem}

\definecolor{amethyst}{rgb}{0.6, 0.4, 0.8}

\usepackage{hyperref}
\hypersetup{
    colorlinks=true,
}

\usepackage[accepted]{icml2024}

\usepackage{amsmath}
\usepackage{amssymb}
\usepackage{mathtools}
\usepackage{amsthm}
\usepackage{listings} %

\usepackage[capitalize,noabbrev,nameinlink]{cleveref}

\usepackage{inconsolata}

\theoremstyle{plain}

\theoremstyle{definition}

\theoremstyle{remark}

\usepackage[textsize=tiny]{todonotes}

\usepackage[loose]{minitoc}

\doparttoc %
\faketableofcontents %

\icmltitlerunning{{\color{Maroon}L}LM Maybe {\color{Maroon}Long}LM: SelfExtend LLM Context Window Without Tuning}
\begin{document}
\normalem
\twocolumn[

\icmltitle{{\color{Maroon}L}LM Maybe {\color{Maroon}Long}LM: SelfExtend LLM Context Window Without Tuning}

\icmlsetsymbol{equal}{*}

\begin{icmlauthorlist}
\icmlauthor{Hongye Jin}{tamu,equal}
\icmlauthor{Xiaotian Han}{tamu,equal}
\icmlauthor{Jingfeng Yang}{amazon}
\icmlauthor{Zhimeng Jiang}{tamu}
\icmlauthor{Zirui Liu}{rice}
\icmlauthor{Chia-Yuan Chang}{tamu}
\icmlauthor{Huiyuan Chen}{wcase}
\icmlauthor{Xia Hu}{rice}
\end{icmlauthorlist}

\icmlaffiliation{tamu}{Texas A\&M University}
\icmlaffiliation{amazon}{Amazon, the views expressed or the conclusions reached are his own and do not represent the view of Amazon}
\icmlaffiliation{wcase}{Case Western Reserve University}
\icmlaffiliation{rice}{Rice University}

\icmlkeywords{Machine Learning, ICML}

\vskip 0.3in
]

\icmlcorrespondingauthor{Hongye Jin}{jhy0410@tamu.edu}
\printAffiliationsAndNotice{\icmlEqualContribution} %

\begin{abstract}

It is well known that LLMs cannot generalize well to long contexts whose lengths are larger than the training sequence length. This poses challenges when employing LLMs for processing long input sequences during inference.  In this work, we argue that LLMs themselves have inherent capabilities to handle long contexts without fine-tuning.  To achieve this goal, we propose SelfExtend to extend the context window of LLMs by constructing bi-level attention information: the grouped attention and the neighbor attention. The grouped attention captures the dependencies among tokens that are far apart, while neighbor attention captures dependencies among adjacent tokens within a specified range.  The two-level attentions are computed based on the original model's self-attention mechanism during inference. With minor code modification, our SelfExtend can effortlessly extend existing LLMs' context window without any fine-tuning. We conduct comprehensive experiments on multiple benchmarks and the results show that our SelfExtend can effectively extend existing LLMs' context window length. The code can be found at \url{https://github.com/datamllab/LongLM}.
\end{abstract}

\section{Introduction}
\label{intro}
The context window length of most existing LLMs~\cite{zhao2023survey,yang2023harnessing} is limited since they are trained with a fixed length of training sequences. It's determined by the context window length during the pretraining stage. Once the length of the input texts exceeds the pretraining context window during the inference, the behavior of LLMs will be unpredictable and suffer from severe performance degradation. 
The perplexity (PPL) of the model will explode with the long input sequences~\cite{xiao2023efficient, peng2023yarn, han2023lm, chen2023extending}. 

Recently, a variety of context window extension methods have been developed to extend the context window of pretrained LLMs. A straightforward approach is to fine-tune these models on enough extensive texts. 
Besides this, some methods seek to extend context window length in more efficient fine-tuning ways. Among these contemporary methods, some notable techniques include `PI'~\cite{chen2023extending}, `CLEX'~\cite{chen2023clex} `Yarn'~\cite{peng2023yarn}, `PoSE'~\cite{zhu2023pose}, `LongLora'~\cite{chen2023longlora}, and `ABF'~\cite{xiong2023effective}. These methods aim to extend the content window based on the implicit assumption that pretrained LLMs \emph{lack the ability to handle long content}.
However, these methods typically require finetuning to achieve extension, which can be resource and time-intensive given the quadratic complexity of Transformers. Additionally, high-quality long text data is scarce, hindering such fine-tuning approaches. Most real-world data is short, and much long text lacks meaningful long-range dependencies. With limited appropriate data, finetuning risks degrading existing strong performance on shorter sequences from pretraining or overfitting models to the tuning set. LLMs' generalizability to broad tasks may be reduced.

Instead of extending the content window, in this paper, we believe \textbf{LLMs should have inherent capabilities to handle long contexts}. Our belief stems from the fact that when we, as human beings, are children, we are taught how to read and write using relatively short texts, such as articles spanning several pages. We rarely use extremely long texts like entire books or complete documents as learning materials. Yet, we are still able to understand long texts effectively. With this strong motivation, the poor performance of LLMs while facing long text is not due to the lack of long context understanding capabilities. 
In our analysis, the key challenge preventing LLMs from effectively handling longer contexts is the Out-of-Distribution (O.O.D) issues related to positional encoding, which we call the \emph{positional O.O.D}\footnote{Here, the position refers to relative position rather than absolute position. The relative position is $m-n$ in RoPE, where $m$ and $n$ are the absolute positions of two tokens. The \emph{positional O.O.D} refers to cases where the value of $m-n$ during inference is unseen, i.e., larger than the values observed during pretraining. In this paper, we map unseen large relative positions to those observed during pretraining. More details about $m-n$ are provided in~\cref{sec:prel}.} issue. 
This problem arises when LLMs encounter text sequences during inference exceeding the length of their pretraining context window, where LLMs are exposed to new relative distances that were not present during their pretraining phase. It is widely recognized that Neural Networks (NNs) are susceptible to unpredictable behaviors when dealing with O.O.D inputs~\cite{liu2021towards, shen2021towards, bai2021recent, zhang2023neural}. To address this, an intuitive and practical solution would be to remap the unseen relative positions to those encountered during the pretraining, thus extending the LLMs' ability to handle longer contexts naturally.

This paper proposes SelfExtend to elicit LLMs' inherent long context capabilities. SelfExtend addresses the issue of O.O.D. positional information by using a simple floor division operation to map unseen large relative positions to those encountered during pretraining. 
The core idea hinges on the observation that, in long texts, exacting word positions becomes less crucial. The overall meaning and the relative order of information hold greater significance. Just like when answering questions about lengthy texts, we rely on the general location and order, not the specific word-by-word placement. Natural language exhibits a characteristic where meaning stays relatively consistent within short ranges like paragraphs. Therefore, using close or even identical position encodings effectively captures the necessary relative ordering of important information. This intuitive approach aligns perfectly with the floor operation's functionality. Additionally, T5~\cite{raffel2020exploring} and iRPE~\cite{wu2021rethinking} also share this similar intuition.

Our SelfExtend is a plug-and-play method that takes effect at the inference stage, allowing existing large language models to easily adopt it. We evaluate SelfExtend with some popular LLMs~(Llama-2~\cite{touvron2023llama}, Mistral~\cite{jiang2023mistral}, SOLAR~\cite{kim2023solar}, and Phi-2~\cite{javaheripi2023phi}) on three types of tasks: language modeling, synthetic long context tasks, and real-world long context tasks. The proposed SelfExtend substantially improves the long context understanding ability and even outperforms many finetuning-based methods on some tasks. These results underscore SelfExtend as an effective solution for context window extension. The superior performance of SelfExtend also demonstrated the potential of large language models to effectively handle long contexts. Our main contributions are summarized as follows:
\begin{itemize}[leftmargin=0.4cm, itemindent=.0cm, itemsep=0.0cm, topsep=0.0cm]
    \item We think LLMs with RoPE have a natural ability to handle long texts, even if they have not encountered super-long ones during training. The previous limitation stems from  O.O.D. positions, meaning the "larger" positions have not been seen during training. We call this the \emph{positional O.O.D.} issue.
    \item Based on this belief and to address the positional O.O.D. issue, we propose SelfExtend to extend the context window of LLMs without any fine-tuning. We map the unseen large relative positions (at inference) to known positions (at training), thus allowing LLMs to maintain coherence over longer texts without additional fine-tuning.
    \item In both synthetic and real-world long context tasks, SelfExtend has proven its ability to deliver performance that matches or surprisingly surpasses many existing fine-tuning-based models. This highlights the superior capabilities of our SelfExtend model.
\end{itemize}

\section{Preliminary}\label{sec:prel}
In this section, we present the preliminaries of our work.

\textbf{Position Encoding.} Transformers~\cite{vaswani2017attention} incorporate position information via different positional embedding designs. The common positional embedding design can generally be categorized into two classes: absolute position embeddings and relative positional encodings. The \emph{absolute position embedding} provides the absolute positions, which embeds each absolute position $i$ into position vector $\mathbf{p}_i$ and adds word embeddings to their corresponding $\mathbf{p}_i$ before feeding them to the model. Examples of such include sinusoidal position embeddings~\cite{vaswani2017attention} and learned position embeddings in GPT3~\citep{brown2020language} and OPT~\citep{zhang2022opt}, or adding the dot product between two tokens' position embeddings on the attention logit~\cite{ke2020rethinking}. On the other hand, relative positional encodings have been proposed to use relative distance information between tokens and have become the mainstream of position embedding. This information is usually applied in attention layers. Examples of such include a learnable attention logit bias as in T5~\cite{xue2020mt5}, Transformer-XL~\cite{dai2019transformer}; a fixed linear attention decay called Alibi~\cite{press2021train}; rotating query and key sequences based on distance such as RoPE~\cite{su2022roformer}, and XPos~\cite{xpos}. The proposed method in this work is based on the Rotary Position Embedding (RoPE) introduced in \cite{su2022roformer}.

\textbf{RoPE.} Here, we introduce the basic concept of RoPE. Let's consider a sequence of tokens represented as $w_1, w_2, \cdots, w_L$, and their corresponding embeddings are denoted as $\textbf{x}_1, \cdots, \textbf{x}_L\in \mathbb R^{|D|}$, where $|D|$ is the dimension of the embedding. The basic idea of RoPE is to incorporate the positional information into the query $\mathbf{q}$ and the key vectors $\mathbf{k}$, respectively. This integration ensures that their inner product $\mathbf{q}^T\mathbf{k}$ will contain the relative positional embedding information inherently. To achieve this, RoPE employs the following vector transformations: 
\begin{align}
\mathbf q_m = f_q(\textbf{x}_m, m) \in \mathbb R^{|L|}, ~ \mathbf k_n = f_k(\textbf{x}_n, n) \in \mathbb R^{|L|},
\end{align}
where $|L|$ is the hidden dimension of per head. The functions $f_q$ and  $f_k$ responsible for injecting positional information, are defined as $f_q(\textbf{x}_m, m) = W_q\textbf{x}_me^{im\theta}, ~ f_k(\textbf{x}_n, n) = W_k\textbf{x}_ne^{in\theta},$
where $\theta_d = b^{-2d/|D|}$,  $b=10000$ and  projectors $W_q, W_k: \mathbb R^{|D|}\rightarrow \mathbb R^{|L|}$. RoPE keeps the real part of the inner product $\mathbf q^T\mathbf k$, which is $\text{Re}(\mathbf q^*\mathbf k)$.  This operation ensures that the dot product of the query and key vectors depends entirely on the relative distance between the tokens, represented by $m - n$ of the tokens as follows:
\begin{align}
\nonumber
 & \langle f_q(\textbf{x}_m, m), f_k(\textbf{x}_n, n)\rangle_{\mathbb R} = \text{Re}(\langle f_q(\textbf{x}_m, m), f_k(\textbf{x}_n, n)\rangle_{\mathbb C})\\
=& \text{Re}(\textbf{x}_m^* W_q^*W_k\textbf{x}_ne^{i\theta(m - n)}) = g(\textbf{x}_m, \textbf{x}_n, m - n),
\end{align}
where $g(\cdot)$ is an abstract mapping function.

\section{SelfExtend}\label{sec:method}

In this section, we first conduct a preliminary investigation on the inherent ability of the LLMs to handle long content. Then, we propose our SelfExtend that effectively extends existing LLMs' context window without any fine-tuning.

\subsection{Preliminary Analysis}

\textbf{\ding{172} Why do LLMs fail on sequences during inference that are longer than their pre-training context window?} For a pretrained LLM with relative position encodings, such as RoPE, the behavior of the LLMs becomes unpredictable during inference if the length of a sequence is longer than its pretraining context window length. This has been explored by \cite{han2023lm, chen2023extending} that with unseen relative positions, the attention distributions are different compared to those within the pretraining context window. We argue that such failure stems from the Out-of-Distribution (O.O.D.) relative distance in the sense that neural networks are not robust to O.O.D. inputs \cite{shen2021towards}.

\begin{figure}[!t]
    \centering
    \includegraphics[width=\columnwidth]{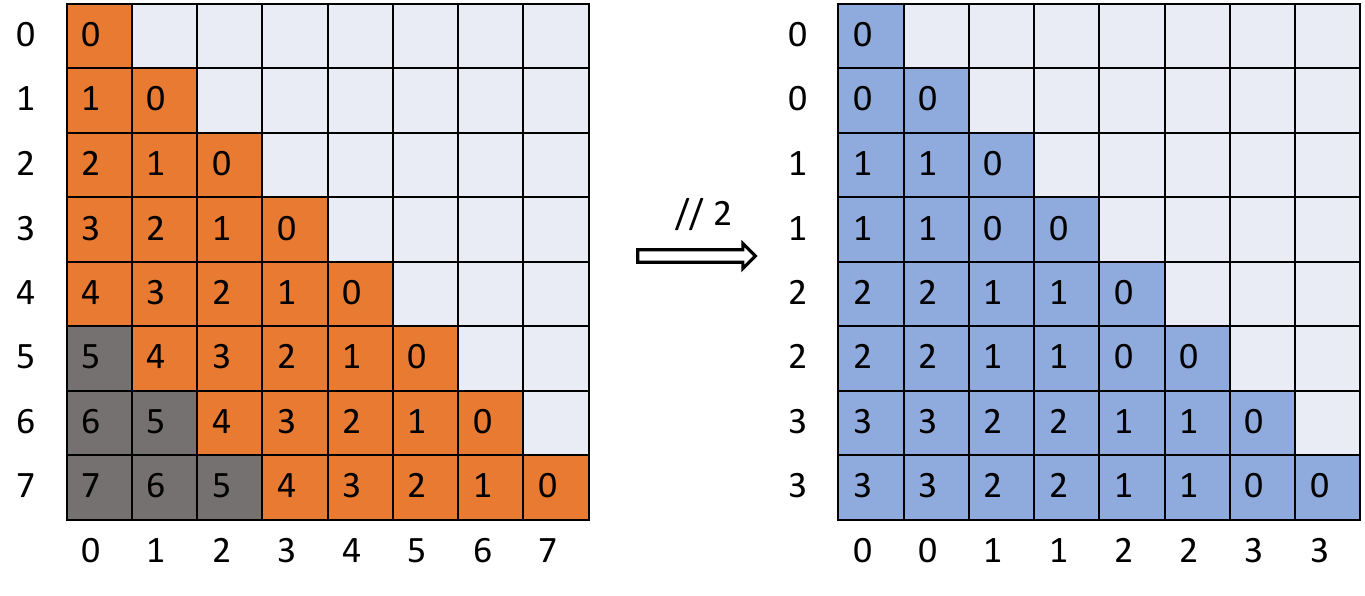}\vspace{-18pt}
    \caption{Illustration of grouped attention. We suppose that the LLM's pretraining context window length is $5$ and the length of the inference sequence is $8$. On the left figure, we show the positional Out-of-Distribution (O.O.D.) issue while the input length is out of the pretraining context window size. The y-axis of this matrix represents the position of query tokens and the x-axis represents the position of key tokens. In this case, in the relative position matrix, only those in \textcolor{orange}{orange} are seen during pretraining. Relative positions in \textcolor{gray}{gray} are outside the pretraining context window. In the right figure, we show how the \textsc{floor} operation is applied and the relative position matrix for grouped self attention. With the $G_s$ set as 2, the positions of query tokens and key tokens are mapped from 0-7 to 0-3 by \textsc{floor} ($//$). The new relative positions (in \textcolor{blue}{blue}) are all within the range of the pretraining context window.} \label{fig:group_attn}
    \vspace{-10pt}
\end{figure}

\textbf{\ding{173} How to solve positional O.O.D. problem?}
One feasible and straightforward way to handle unseen relative positions is to map them to positions that were seen during pretraining. We can use the \textsc{floor} operation to map the unseen positions to positions within the pretraining context window, as shown in Figure~\ref{fig:group_attn}. The proposed method is identical to the original self-attention mechanism except that the \textsc{floor} operation is applied to each token's original position before the inner product. We denote the self attention with the \textsc{floor} operation applied as ``grouped attention''.  In Python style, the ``grouped attention'' is denoted as:
\begin{equation}
    P_g = P~~~//~~~G_s,
\end{equation}
where $P \in \mathbb{R}^{B \times L}$ is the original position encoding, in which $B$ is the batch size and $L$ is the length of the input text sequence. $G_s$ denotes the group size, which is the base of the \textsc{floor} operation. Taking the floor of the position divided by the group size maps the original large position values to a smaller discrete set of values, avoiding the issue of out-of-distribution position values during inference.

\begin{figure}[t]
    \centering
    \includegraphics[width=0.9\linewidth]{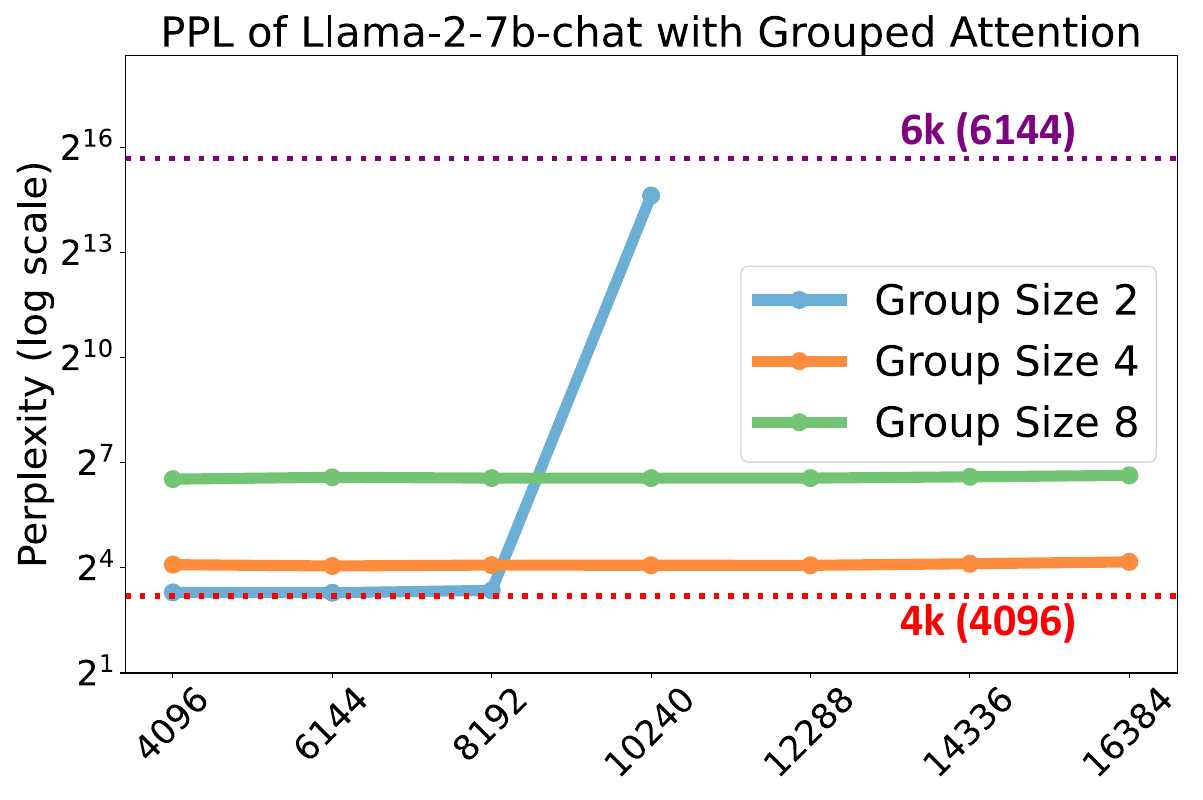}
    \vspace{-15pt}
    \caption{Perplexity~(PPL) using grouped attention with different group sizes under different sequence lengths on PG-19 dataset. The original Llama-2-7b-chat PPL is stable at {\color{red} 4k (4096)} sequences (red dotted line) but explodes at {\color{amethyst} 6k (6144)} sequences (purple dotted line). The results show the LLMs keep a relatively low and stable PPL on long sequences with grouped attention.}
    \label{fig:group_ppl}
    \vspace{-10pt}
\end{figure}

\textbf{\ding{174} Can LLMs work well without accurate position information? --- Yes, but not that perfect.} We show the perplexity~(PPL) on the PG-19~\cite{rae2019compressive} dataset with the \textsc{floor} operation applied to Llama-2-7b-chat across different sequence lengths, in \cref{fig:group_ppl}. As a comparison, we also show the PPL of the original model without the \textsc{floor} operation as the dotted lines. From this figure, with the \textsc{floor} operation, LLMs keep a relatively low and stable PPL on the sequences whose lengths exceed the pretraining context window. Meanwhile, with grouped attention, the PPL is a little higher than the original LLMs, which is expected. However, the model's PPL behavior is similar to the original model, as the PPL is nearly unchanged within the ``context window'' (for Llama-2: 2 - 8192, 4 - 16384, and 8 - 32768), demonstrating the effectiveness of group attention. Once the length of a sequence is longer than the new ``context window''~(e.g., sequences with 10k tokens as the input, with a group size of 2 ), the PPL explodes again due to the \emph{positional O.O.D} issue.

\begin{figure*}[t]
    \centering
    \includegraphics[width=0.85\textwidth]{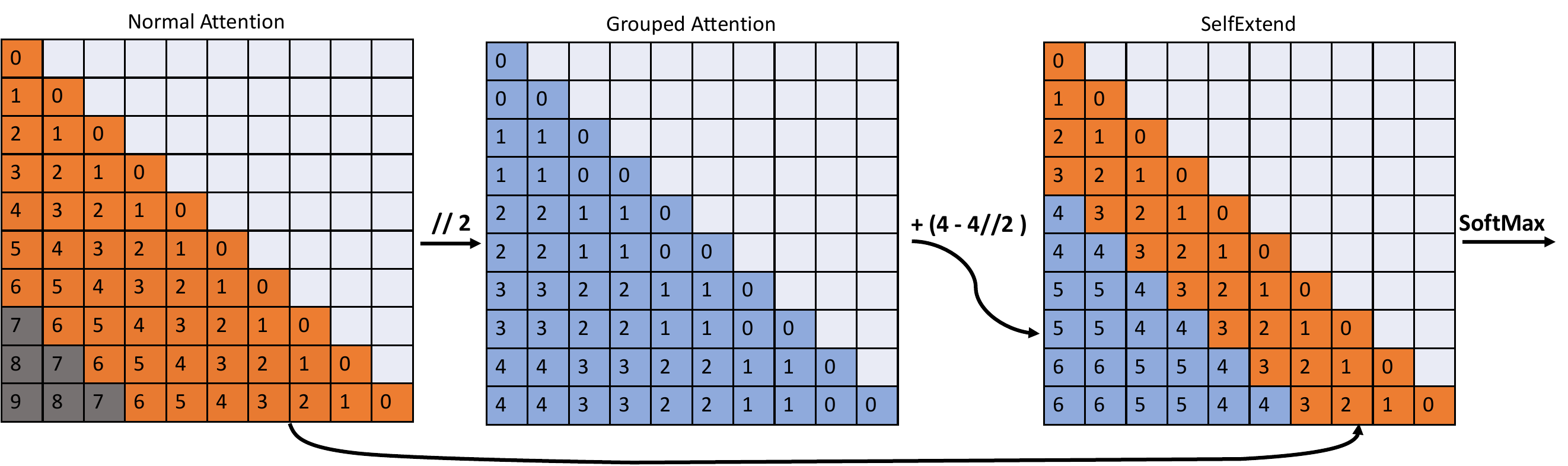}\vspace{-10pt}
    \caption{Illurstation of SelfExtend. This figure shows the attention score matrix~(before SoftMax operation) of SelfExtend while a sequence of length $10$ is fed into an LLM with the pretraining context window size~($L=7$). The numbers denote the relative distances between the corresponding query and key tokens. SelfExtend has two kinds of attention mechanism: for neighbor tokens within the neighbor window~($w_n=4$), it adapts the normal self-attention; for tokens out of the window, it adapts the values from the grouped attention. The group size~($G_s$) is set to 2. We then merge two parts attention matrices and apply the softmax operation. }
    \label{fig:self_ext}
    \vspace{-10pt}
\end{figure*}

\textbf{\ding{175} How to restore degraded language modeling ability caused by grouped attention? --- Re-introducing normal attention in the neighboring area.}
In the process of generating next tokens, the immediate neighbors of a target token play a crucial role, which is well-supported by existing methods of sparse attention mechanisms~\cite{zaheer2020big, shi2021sparsebert} and methods for extending the contextual window~\cite{han2023lm, xiong2023effective,liu2024kivi}. These studies consistently highlight the importance of maintaining the standard attention mechanism for tokens in close proximity to the target token. This proximity-based focus is essential for the accurate generation of the next token, ensuring the coherence and fluency of the generated text, as evidenced by acceptable perplexity (PPL) levels. Employing grouped attention might not significantly affect the overall quality of generated sentences; however, it necessitates the accurate positioning of attention to maintain generation quality. Therefore, it is imperative to preserve the standard attention mechanism within the vicinity of the target token, as utilized during the pretraining phase, to ensure the precision and effectiveness of language models in capturing the nuances of local context.

\subsection{SelfExtend LLM Context Window Without Tuning}
We introduce SelfExtend, a method that enhances LLMs' natural capability to process extensive contexts without the need for fine-tuning. SelfExtend incorporates two distinct types of attention mechanisms: 1) Grouped attention, specifically designed for tokens that are far apart. This approach applies a floor operation to the positions to manage long-distance relationships between tokens; 2) Standard attention, which employs the conventional attention mechanism for adjacent tokens within a specified range. The SelfExtend framework is depicted in \cref{fig:self_ext}. Notably, SelfExtend modifies only the attention mechanism during inference, eliminating the need for additional fine-tuning.

\textbf{Maximum Extended Length of SelfExtend} Suppose that we have the pretraining context window size as $L$, the group size for grouped attention as $G_s$, and the window size for neighbor tokens as $w_n$. We shift the relative position of grouped attention by $w_n - w_n // G_s$ before merging the two pieces of attention together. This ensures that the transition from the normal attention area to the grouped attention area smooth. We merge the two parts of attention by replacing the attention values out of the neighbor token window with the attention values from the grouped attention. All the modifications are applied before the softmax operation and other parts remain unchanged. Ideally, the maximum length of the extended context window is:
\begin{equation}\label{eq:length}
    (L - w_n) * G_s + w_n.
\end{equation}
For example, in \cref{fig:self_ext}, the context window is extended from its pretraining length of $7$ to $(7 - 4) * 2 + 4 = 10$. The pseudo code for SelfExtend is presented in \cref{alg:SelfExtend}.

\textbf{Relation to Existing Work} The grouped attention in SelfExtend can be viewed as a form of position interpolation~\cite{chen2023extending}, where some positions are interpolated to be infinitely close to pretraining positions. Another finetuning-free method, ReRoPE~\cite{rerope2023}, is equivalent to a special case of SelfExtend: the group size is large enough that all tokens outside the neighbor window fall into the same group (e.g. group size $10,000$ in~\cref{fig:hype}). T5~\cite{raffel2020exploring} and iRPE~\cite{wu2021rethinking} also share the high-level idea of multi-level positional encodings, while applying it during pretraining. T5 is more similar to ReRoPE for using the same position for distant tokens. iRPE has finer distant position encodings, more akin to SelfExtend.

\begin{table}[t]
\centering
\caption{Perplexity on dataset PG19 with Llama-2-7b-chat and Mistral-7b-instruct-0.1. We report the PPL of with\&without Sliding Window Attention~(SWA) for Mistral.}\label{table:ppl}

\resizebox{1.01\linewidth}{!}{
    \begin{tabular}{@{}lcccccccc@{}}
    \toprule
    Model & \multicolumn{6}{c}{Evaluation Context Window Size} \\ 
    \textbf{Name}   & \textbf{4096} &\textbf{6144} & \textbf{8192} & \textbf{10240} & \textbf{12288} &  \textbf{14336} & \textbf{16384} \\
    \midrule
    Llama-2-7b-chat & 9.181 &  $>10^3$ & $>10^3$ & $>10^3$ & $>10^3$ & $>10^3$ &$> 10^3$  \\
    SelfExtend-Llama-2-7b-chat  & 8.885 & 8.828 & 9.220 & 8.956 & 9.217 & 9.413 & 9.274 \\
    \midrule
    Mistral-7b-instruct-0.1 w/ SWA  & 9.295 & 9.197 & 9.532 & 9.242 & 9.198 & 9.278 & 9.294\\
    Mistral-7b-instruct-0.1 w/o SWA  & 9.295& 9.205 & 10.20 & 55.35 & $> 10^3$ & $> 10^3$ & $> 10^3$ \\
    SelfExtend-Mistral-7b-instruct-0.1  & 9.272 & 9.103 & 9.369 & 9.070 & 8.956 & 9.022 & 9.128  \\
    \bottomrule
    \end{tabular}
}\vspace{-10pt}
\end{table}

\section{Experiments}

We evaluate SelfExtend with Llama-2~\cite{touvron2023llama} and its families, Phi-2~\cite{javaheripi2023phi}, Mistral~\cite{jiang2023mistral} and SOLAR~\cite{kim2023solar} on language modeling task, synthetic long context tasks, real-world long context tasks and standard short-context tasks.

\begin{table*}[t]

\fontsize{18}{24}\selectfont
\setlength{\tabcolsep}{5pt}
\centering
\caption{Performance comparison of different LLMs on LongBench. {\color{blue}*} indicates the results reported by LongBench. {\color{yellow} *}indicates the results are reported by CLEX~\cite{chen2023clex}. {\color{red} +} indicates the results from us. Models in green/blue/cyan/orange are based on Llama2-7b/Mistral-7b/Phi-2/SOLAR-10.5B. The number~(e.g. `25k') indicates the maximum input length. The `SE' prefix indicates SelfExtend is applied to this model. In this table, except SelfExtend, all other models require fine-tuning to extend the context window. CLEX is fine-tuned with 2B tokens. LongChat1.5-7B-32k and Vicuna1.5-7B-16K are fine-tuned on more than 80k conversations. CodeLLaMA~\cite{roziere2023code} is fine-tuned on more than 500B tokens. MistralLite~\cite{MistralLite_2023} is also fine-tuned on more than 2B tokens~\cite{amazon2023mistrallite}. The better performance between models w/ and w/o SelfExtend is in \textbf{bold}.}\label{tab:longbench}

\begin{threeparttable}

\scalebox{0.3}{
\begin{tabular}{l|lcccccccccccccccc}
\specialrule{1pt}{0pt}{2pt}
&\multirow{4}{*}{~~~~~~~~~~~~~~~~~~~~LLMs\tnote{a}} & \multicolumn{3}{c}{Single-Document QA} & \multicolumn{3}{c}{Multi-Document QA}& \multicolumn{3}{c}{Summarization}& \multicolumn{3}{c}{Few-shot Learning}& \multicolumn{2}{c}{Synthetic} & \multicolumn{2}{c}{Code} \\
\cmidrule(lr){3-5}\cmidrule(lr){6-8}\cmidrule(lr){9-11}\cmidrule(lr){12-14}\cmidrule(lr){15-16}\cmidrule(lr){17-18}
&& \rotatebox[origin=c]{30}{NarrativeQA} & \rotatebox[origin=c]{30}{Qasper} & \rotatebox[origin=c]{30}{MultiField-en} & \rotatebox[origin=c]{30}{HotpotQA} & \rotatebox[origin=c]{30}{2WikiMQA} & \rotatebox[origin=c]{30}{Musique} & \rotatebox[origin=c]{30}{GovReport} & \rotatebox[origin=c]{30}{QMSum} & \rotatebox[origin=c]{30}{MultiNews} & \rotatebox[origin=c]{30}{TREC} & \rotatebox[origin=c]{30}{TriviaQA} & \rotatebox[origin=c]{30}{SAMSum} & \rotatebox[origin=c]{30}{PassageCount} & \rotatebox[origin=c]{30}{PassageRe} & \rotatebox[origin=c]{30}{Lcc} & \rotatebox[origin=c]{30}{RepoBench-P} \\

\specialrule{1pt}{2pt}{2pt}

\multirow{13}{*}{\rotatebox[origin=c]{90}{\fontsize{18}{100}\selectfont SelfExtend}}&\cellcolor{green!10}~~~~~~Llama-2-7B-chat-4k{\color{blue}*} & \cellcolor{green!10}18.7&\cellcolor{green!10}19.2&\cellcolor{green!10} \textbf{36.8} &\cellcolor{green!10} 25.4 &\cellcolor{green!10} 32.8&\cellcolor{green!10} 9.4 &\cellcolor{green!10} 27.3 & \cellcolor{green!10}20.8 &\cellcolor{green!10} 25.8 &\cellcolor{green!10} 61.5 &\cellcolor{green!10} 77.8 & \cellcolor{green!10} 40.7 &\cellcolor{green!10} 2.1 &\cellcolor{green!10}\textbf{9.8} & \cellcolor{green!10}52.4 & \cellcolor{green!10}43.8\\
&\cellcolor{green!10}SE-Llama-2-7B-chat-16k{\color{red} +}&\cellcolor{green!10}\textbf{21.69}&\cellcolor{green!10}25.02&\cellcolor{green!10}35.21&\cellcolor{green!10}34.34&\cellcolor{green!10}30.24            &\cellcolor{green!10}14.13         &\cellcolor{green!10}27.32&\cellcolor{green!10}\textbf{21.35}&\cellcolor{green!10}25.78&\cellcolor{green!10}\textbf{69.50} &\cellcolor{green!10}\textbf{81.99}&\cellcolor{green!10}40.96&\cellcolor{green!10}\textbf{5.66}&\cellcolor{green!10}5.83&\cellcolor{green!10}\textbf{60.60}&\cellcolor{green!10}\textbf{54.33}\\ 
&\cellcolor{green!10}SE-Llama-2-7B-chat-25k{\color{red} +}&\cellcolor{green!10}21.37&\cellcolor{green!10}\textbf{26.68}&\cellcolor{green!10}34.63&\cellcolor{green!10}\textbf{35.47}   &\cellcolor{green!10}\textbf{30.46}&\cellcolor{green!10}\textbf{15.51}&\cellcolor{green!10}\textbf{27.51}&\cellcolor{green!10}21.30&\cellcolor{green!10}\textbf{25.87}&\cellcolor{green!10}68.50&\cellcolor{green!10}78.79&\cellcolor{green!10}\textbf{41.29}&\cellcolor{green!10}3.90            &\cellcolor{green!10}3.50&\cellcolor{green!10}59.69            &\cellcolor{green!10}53.83\\ \cline{2-18}

& \cellcolor{blue!10}~~~~~~Mistral-7B-ins-0.1-16k w/ SWA{\color{red} +} & \cellcolor{blue!10}19.40 & \cellcolor{blue!10}34.53 & \cellcolor{blue!10}37.06 & \cellcolor{blue!10}42.29 & \cellcolor{blue!10}32.49 & \cellcolor{blue!10}14.87 & \cellcolor{blue!10}27.38 & \cellcolor{blue!10}22.75 & \cellcolor{blue!10}26.82 & \cellcolor{blue!10}65.00 & \cellcolor{blue!10}\textbf{87.77} & \cellcolor{blue!10}42.34 & \cellcolor{blue!10}\textbf{1.41} & \cellcolor{blue!10}28.50 & \cellcolor{blue!10}\textbf{57.28} & \cellcolor{blue!10}53.44 \\
& \cellcolor{blue!10}~~~~~~Mistral-7B-ins-0.1-8k w/o SWA{\color{red} +} & \cellcolor{blue!10}20.46 & \cellcolor{blue!10}35.36 & \cellcolor{blue!10}39.39 & \cellcolor{blue!10}34.81 & \cellcolor{blue!10}29.91 & \cellcolor{blue!10}11.21 & \cellcolor{blue!10}24.70 & \cellcolor{blue!10}21.67 & \cellcolor{blue!10}26.67 & \cellcolor{blue!10}68.00 & \cellcolor{blue!10}86.66 & \cellcolor{blue!10}41.28 & \cellcolor{blue!10}0.18 & \cellcolor{blue!10}24.00 & \cellcolor{blue!10}56.94 & \cellcolor{blue!10}\textbf{55.85} \\ 
& \cellcolor{blue!10}SE-Mistral-7B-ins-0.1-16k{\color{red} +}\tnote{b} & \cellcolor{blue!10}\textbf{23.56} & \cellcolor{blue!10}\textbf{39.33} & \cellcolor{blue!10}\textbf{49.50} & \cellcolor{blue!10}\textbf{45.28} & \cellcolor{blue!10}\textbf{34.92} & \cellcolor{blue!10}\textbf{23.14} & \cellcolor{blue!10}\textbf{30.71} & \cellcolor{blue!10}\textbf{24.87} & \cellcolor{blue!10}\textbf{26.83} & \cellcolor{blue!10}\textbf{69.50} & \cellcolor{blue!10}86.47 & \cellcolor{blue!10}\textbf{44.28} & \cellcolor{blue!10}1.18 & \cellcolor{blue!10}\textbf{29.50} & \cellcolor{blue!10}55.32 & \cellcolor{blue!10}53.44 \\ 
\cline{2-18}

& \cellcolor{cyan!10}~~~~~~Phi-2-2k{\color{red} +} & \cellcolor{cyan!10}4.46 & \cellcolor{cyan!10}7.01 & \cellcolor{cyan!10}19.98 & \cellcolor{cyan!10}\textbf{9.43} & \cellcolor{cyan!10}8.55 & \cellcolor{cyan!10}\textbf{4.62} & \cellcolor{cyan!10}25.64 & \cellcolor{cyan!10}14.32 & \cellcolor{cyan!10}\textbf{24.03} & \cellcolor{cyan!10}50.50 & \cellcolor{cyan!10}74.55 & \cellcolor{cyan!10}\textbf{1.71} & \cellcolor{cyan!10}\textbf{2.83} & \cellcolor{cyan!10}\textbf{4.17} & \cellcolor{cyan!10}\textbf{58.96} & \cellcolor{cyan!10}54.14 \\
& \cellcolor{cyan!10}SE-Phi-2-8k{\color{red} +} & \cellcolor{cyan!10}\textbf{12.04} & \cellcolor{cyan!10}\textbf{12.10} & \cellcolor{cyan!10}\textbf{20.15} & \cellcolor{cyan!10}8.22 & \cellcolor{cyan!10}\textbf{9.68} & \cellcolor{cyan!10}3.89 & \cellcolor{cyan!10}\textbf{27.90} & \cellcolor{cyan!10}\textbf{14.58} & \cellcolor{cyan!10}22.13 & \cellcolor{cyan!10}\textbf{61.00} & \cellcolor{cyan!10}\textbf{82.82} & \cellcolor{cyan!10}1.40 & \cellcolor{cyan!10}2.37 & \cellcolor{cyan!10}2.83 & \cellcolor{cyan!10}57.87 & \cellcolor{cyan!10}\textbf{56.42} \\
\cline{2-18}

& \cellcolor{red!10}~~~~~~SOLAR-10.7B-ins-4k{\color{red} +} & \cellcolor{red!10}16.50 & \cellcolor{red!10}24.06 & \cellcolor{red!10}46.76 & \cellcolor{red!10}44.03 & \cellcolor{red!10}\textbf{36.05} & \cellcolor{red!10}22.76 & \cellcolor{red!10}\textbf{31.39} & \cellcolor{red!10}19.81 & \cellcolor{red!10}\textbf{26.36} & \cellcolor{red!10}70.00 & \cellcolor{red!10}87.91 & \cellcolor{red!10}42.49 & \cellcolor{red!10}\textbf{4.5} & \cellcolor{red!10}26.5 & \cellcolor{red!10}41.04 & \cellcolor{red!10}54.36 \\
& \cellcolor{red!10}SE-SOLAR-10.7B-ins-16k{\color{red} +} & \cellcolor{red!10}\textbf{22.63} & \cellcolor{red!10}\textbf{32.49} & \cellcolor{red!10}\textbf{47.88} & \cellcolor{red!10}\textbf{46.19} & \cellcolor{red!10}34.32 & \cellcolor{red!10}\textbf{27.88} & \cellcolor{red!10}30.75 & \cellcolor{red!10}\textbf{22.10} & \cellcolor{red!10}25.62 & \cellcolor{red!10}\textbf{74.50} & \cellcolor{red!10}\textbf{89.04} & \cellcolor{red!10}\textbf{42.79} & \cellcolor{red!10}4.0 & \cellcolor{red!10}\textbf{28.0} & \cellcolor{red!10}\textbf{53.73} & \cellcolor{red!10}\textbf{56.47} \\
\cline{2-18}

& \cellcolor{brown!10}~~~~~~Llama-3-8B-ins-8k{\color{red} +} & \cellcolor{brown!10} 21.71& \cellcolor{brown!10}44.24 & \cellcolor{brown!10}44.54 & \cellcolor{brown!10}46.82 & \cellcolor{brown!10}36.42 & \cellcolor{brown!10}21.49 & \cellcolor{brown!10}30.03 & \cellcolor{brown!10}22.67 & \cellcolor{brown!10}\textbf{27.79} & \cellcolor{brown!10}74.5 & \cellcolor{brown!10}90.23 & \cellcolor{brown!10}\textbf{42.53} & \cellcolor{brown!10} NA & \cellcolor{brown!10} 67.00 & \cellcolor{brown!10}\textbf{57.00} & \cellcolor{brown!10} 51.22 \\
& \cellcolor{brown!10}SE-Llama-3-8B-ins-16k{\color{red} +} & \cellcolor{brown!10}\textbf{12.04} & \cellcolor{brown!10}\textbf{12.10} & \cellcolor{brown!10}\textbf{20.15} & \cellcolor{brown!10}8.22 & \cellcolor{brown!10}\textbf{9.68} & \cellcolor{brown!10}3.89 & \cellcolor{brown!10}\textbf{27.90} & \cellcolor{brown!10}\textbf{14.58} & \cellcolor{brown!10}22.13 & \cellcolor{brown!10}\textbf{61.00} & \cellcolor{brown!10}\textbf{82.82} & \cellcolor{brown!10}1.40 & \cellcolor{brown!10}2.37 & \cellcolor{brown!10}2.83 & \cellcolor{brown!10}57.87 & \cellcolor{brown!10}\textbf{56.42} \\
& \cellcolor{brown!10}SE-Llama-3-8B-ins-32k~~~~10/96{\color{red} +} & \cellcolor{brown!10}21.50 & \cellcolor{brown!10}43.96 & \cellcolor{brown!10}50.26 & \cellcolor{brown!10}48.18 & \cellcolor{brown!10}28.18 & \cellcolor{brown!10}25.58 & \cellcolor{brown!10}34.88 & \cellcolor{brown!10}23.83 & \cellcolor{brown!10}26.96 & \cellcolor{brown!10}75.50 & \cellcolor{brown!10}88.26 & \cellcolor{brown!10}42.01 & \cellcolor{brown!10}4.12 & \cellcolor{brown!10}88.0 & \cellcolor{brown!10}36.58 & \cellcolor{brown!10}37.73 \\

\specialrule{1pt}{2pt}{10pt}\specialrule{1pt}{2pt}{2pt}

\multirow{14}{*}{\rotatebox[origin=c]{90}{\fontsize{18}{100}\selectfont Other Methods}} & \cellcolor{green!10}LongChat1.5-7B-32k{\color{blue}*} & \cellcolor{green!10}16.9 & \cellcolor{green!10}27.7 & \cellcolor{green!10}41.4 & \cellcolor{green!10}31.5 & \cellcolor{green!10}20.6 & \cellcolor{green!10}9.7 & \cellcolor{green!10}30.8 & \cellcolor{green!10}22.7 & \cellcolor{green!10}26.4 & \cellcolor{green!10}63.5 & \cellcolor{green!10}82.3 & \cellcolor{green!10}34.2 & \cellcolor{green!10}1.0 & \cellcolor{green!10}30.5 & \cellcolor{green!10}53.0 & \cellcolor{green!10}55.3 \\
& \cellcolor{green!10}together/llama-2-7b-32k{\color{red} +} & \cellcolor{green!10} 15.65 & \cellcolor{green!10}10.49 & \cellcolor{green!10}33.43 & \cellcolor{green!10}12.36 & \cellcolor{green!10}12.53 & \cellcolor{green!10}6.19 & \cellcolor{green!10}29.28 & \cellcolor{green!10} 17.18 & \cellcolor{green!10}22.12 & \cellcolor{green!10}71.0 & \cellcolor{green!10}87.79 & \cellcolor{green!10} 43.78 & \cellcolor{green!10}1.0 & \cellcolor{green!10} 23.0 & \cellcolor{green!10} 63.79 & \cellcolor{green!10}61.77 \\			
& \cellcolor{green!10}CLEX-7B-16k{\color{yellow}*} & \cellcolor{green!10}18.05 & \cellcolor{green!10}23.68 & \cellcolor{green!10}44.62 & \cellcolor{green!10}28.44 & \cellcolor{green!10}19.53 & \cellcolor{green!10}9.15 & \cellcolor{green!10}32.52 & \cellcolor{green!10}22.9 & \cellcolor{green!10}25.55 & \cellcolor{green!10}68 & \cellcolor{green!10}84.92 & \cellcolor{green!10}42.82 & \cellcolor{green!10}0 & \cellcolor{green!10}11.5 & \cellcolor{green!10}59.01 & \cellcolor{green!10}56.87 \\
& \cellcolor{green!10}CodeLLaMA-7B-16k{\color{yellow}*} & \cellcolor{green!10}22.93 & \cellcolor{green!10}30.69 & \cellcolor{green!10}43.37 & \cellcolor{green!10}33.05 & \cellcolor{green!10}27.93 & \cellcolor{green!10}14.2 & \cellcolor{green!10}28.43 & \cellcolor{green!10}24.18 & \cellcolor{green!10}26.84 & \cellcolor{green!10}70 & \cellcolor{green!10}84.97 & \cellcolor{green!10}43.43 & \cellcolor{green!10}2 & \cellcolor{green!10}13.5 & \cellcolor{green!10}64.35 & \cellcolor{green!10}55.87 \\
& \cellcolor{green!10}SE-Llama-2-7B-chat-16k{\color{red} +} & \cellcolor{green!10}21.69 & \cellcolor{green!10}25.02 & \cellcolor{green!10}35.21 & \cellcolor{green!10}34.34 & \cellcolor{green!10}30.24 & \cellcolor{green!10}14.13 & \cellcolor{green!10}27.32 & \cellcolor{green!10}21.35 & \cellcolor{green!10}25.78 & \cellcolor{green!10}69.50 & \cellcolor{green!10}81.99 & \cellcolor{green!10}40.96 & \cellcolor{green!10}5.66 & \cellcolor{green!10}5.83 & \cellcolor{green!10}60.60 & \cellcolor{green!10}54.33 \\
& \cellcolor{green!10}SE-Llama-2-7B-chat-25k{\color{red} +} & \cellcolor{green!10}21.37 & \cellcolor{green!10}26.68 & \cellcolor{green!10}34.63 & \cellcolor{green!10}35.47 & \cellcolor{green!10}30.46 & \cellcolor{green!10}15.51 & \cellcolor{green!10}27.51 & \cellcolor{green!10}21.30 & \cellcolor{green!10}25.87 & \cellcolor{green!10}68.50 & \cellcolor{green!10}78.79 & \cellcolor{green!10}41.29 & \cellcolor{green!10}3.90 & \cellcolor{green!10}3.50 & \cellcolor{green!10}59.69 & \cellcolor{green!10}53.83 \\
\cline{2-18}

&\cellcolor{yellow!10}~~~~~~Vicuna1.5-7B-16k{\color{blue}*} &\cellcolor{yellow!10}19.4              &\cellcolor{yellow!10}26.1              &\cellcolor{yellow!10}38.5   &\cellcolor{yellow!10}25.3   &\cellcolor{yellow!10}20.8   &\cellcolor{yellow!10} 9.8   &\cellcolor{yellow!10} 27.9  &\cellcolor{yellow!10}22.8   &\cellcolor{yellow!10}27.2   &\cellcolor{yellow!10}71.5 &\cellcolor{yellow!10}86.2 &\cellcolor{yellow!10} 40.8 &\cellcolor{yellow!10}6.5 &\cellcolor{yellow!10} 4.5 &\cellcolor{yellow!10} 51.0 &\cellcolor{yellow!10} 43.5\\
&\cellcolor{yellow!10}SE-Vicuna1.5-7B-16k{\color{red} +}      &\cellcolor{yellow!10}21.88             &\cellcolor{yellow!10}35.16    &\cellcolor{yellow!10}42.00  &\cellcolor{yellow!10}31.14  &\cellcolor{yellow!10}22.51  &\cellcolor{yellow!10}13.33  &\cellcolor{yellow!10}28.47  &\cellcolor{yellow!10}22.24  &\cellcolor{yellow!10}26.70           &\cellcolor{yellow!10}69.50            &\cellcolor{yellow!10}86.31   &\cellcolor{yellow!10}40.54 &\cellcolor{yellow!10}3.56         &\cellcolor{yellow!10}7.50 &\cellcolor{yellow!10}60.16 &\cellcolor{yellow!10}44.07\\
&\cellcolor{yellow!10}SE-Vicuna1.5-7B-25k{\color{red} +}      &\cellcolor{yellow!10}22.46 &\cellcolor{yellow!10}34.42 &\cellcolor{yellow!10}42.58  &\cellcolor{yellow!10}30.95  &\cellcolor{yellow!10}24.33  &\cellcolor{yellow!10}12.72  &\cellcolor{yellow!10}27.75  &\cellcolor{yellow!10}22.26  &\cellcolor{yellow!10}27.21  &\cellcolor{yellow!10}72.00   &\cellcolor{yellow!10}84.02            &\cellcolor{yellow!10}40.38 &\cellcolor{yellow!10}3.01         &\cellcolor{yellow!10}7.00 &\cellcolor{yellow!10}58.86 &\cellcolor{yellow!10}43.86\\ \cline{2-18}

& \cellcolor{blue!10}MistralLite-16k{\color{red} +} & \cellcolor{blue!10}32.12 & \cellcolor{blue!10}47.02 & \cellcolor{blue!10}44.95 & \cellcolor{blue!10}58.5 & \cellcolor{blue!10}47.24 & \cellcolor{blue!10}31.32 & \cellcolor{blue!10}33.22 & \cellcolor{blue!10}26.8 & \cellcolor{blue!10}24.58 & \cellcolor{blue!10}71.5 & \cellcolor{blue!10}90.63 & \cellcolor{blue!10}37.36 & \cellcolor{blue!10}3 & \cellcolor{blue!10}54.5 & \cellcolor{blue!10}66.27 & \cellcolor{blue!10}65.29 \\
& \cellcolor{blue!10}SE-Mistral-7B-ins-0.1-16k{\color{red} +} & \cellcolor{blue!10}23.85 & \cellcolor{blue!10}37.75 & \cellcolor{blue!10}46.93 & \cellcolor{blue!10}45.35 & \cellcolor{blue!10}34.54 & \cellcolor{blue!10}23.28 & \cellcolor{blue!10}30.45 & \cellcolor{blue!10}23.58 & \cellcolor{blue!10}26.94 & \cellcolor{blue!10}69.50 & \cellcolor{blue!10}85.72 & \cellcolor{blue!10}43.88 & \cellcolor{blue!10}0.59 & \cellcolor{blue!10}28.50 & \cellcolor{blue!10}54.92 & \cellcolor{blue!10}53.44 \\

\cline{2-18}

& \cellcolor{brown!10}Gradient-Llama-3-8B-Inst-262k(32k){\color{red} +} & \cellcolor{brown!10} 21.71& \cellcolor{brown!10}44.24 & \cellcolor{brown!10}44.54 & \cellcolor{brown!10}46.82 & \cellcolor{brown!10}36.42 & \cellcolor{brown!10}21.49 & \cellcolor{brown!10}30.03 & \cellcolor{brown!10}22.67 & \cellcolor{brown!10}\textbf{27.79} & \cellcolor{brown!10}74.5 & \cellcolor{brown!10}90.23 & \cellcolor{brown!10}\textbf{42.53} & \cellcolor{brown!10} NA & \cellcolor{brown!10} 67.00 & \cellcolor{brown!10}\textbf{57.00} & \cellcolor{brown!10} 51.22 \\
& \cellcolor{brown!10}Gradient-Llama-3-8B-Inst-1M(32k){\color{red} +} & \cellcolor{brown!10}\textbf{12.04} & \cellcolor{brown!10}\textbf{12.10} & \cellcolor{brown!10}\textbf{20.15} & \cellcolor{brown!10}8.22 & \cellcolor{brown!10}\textbf{9.68} & \cellcolor{brown!10}3.89 & \cellcolor{brown!10}\textbf{27.90} & \cellcolor{brown!10}\textbf{14.58} & \cellcolor{brown!10}22.13 & \cellcolor{brown!10}\textbf{61.00} & \cellcolor{brown!10}\textbf{82.82} & \cellcolor{brown!10}1.40 & \cellcolor{brown!10}2.37 & \cellcolor{brown!10}2.83 & \cellcolor{brown!10}57.87 & \cellcolor{brown!10}\textbf{56.42} \\
& \cellcolor{brown!10}SE-Llama-3-8B-ins-32k~~~~10/96{\color{red} +} & \cellcolor{brown!10}\textbf{12.04} & \cellcolor{brown!10}\textbf{12.10} & \cellcolor{brown!10}\textbf{20.15} & \cellcolor{brown!10}8.22 & \cellcolor{brown!10}\textbf{9.68} & \cellcolor{brown!10}3.89 & \cellcolor{brown!10}\textbf{27.90} & \cellcolor{brown!10}\textbf{14.58} & \cellcolor{brown!10}22.13 & \cellcolor{brown!10}\textbf{61.00} & \cellcolor{brown!10}\textbf{82.82} & \cellcolor{brown!10}1.40 & \cellcolor{brown!10}2.37 & \cellcolor{brown!10}2.83 & \cellcolor{brown!10}57.87 & \cellcolor{brown!10}\textbf{56.42} \\

\specialrule{1pt}{2pt}{10pt} \specialrule{1pt}{2pt}{2pt}

\multirow{7}{*}{\rotatebox[origin=c]{90}{\fontsize{18}{100}\selectfont Fixed Models}} & \cellcolor{gray!10}GPT-3.5-Turbo-16k{\color{blue}*} & \cellcolor{gray!10}23.6 & \cellcolor{gray!10}43.3 & \cellcolor{gray!10}52.3 & \cellcolor{gray!10}51.6 & \cellcolor{gray!10}37.7 & \cellcolor{gray!10}26.9 & \cellcolor{gray!10}29.5 & \cellcolor{gray!10}23.4 & \cellcolor{gray!10}26.7 & \cellcolor{gray!10}68.0 & \cellcolor{gray!10}91.4 & \cellcolor{gray!10}41.7 & \cellcolor{gray!10}4.5 & \cellcolor{gray!10}71.0 & \cellcolor{gray!10}54.7 & \cellcolor{gray!10}53.6 \\
& \cellcolor{gray!10}XGen-7B-8k{\color{blue}*} & \cellcolor{gray!10}18 & \cellcolor{gray!10}18.1 & \cellcolor{gray!10}37.7 & \cellcolor{gray!10}29.7 & \cellcolor{gray!10}21.1 & \cellcolor{gray!10}10.3 & \cellcolor{gray!10}27.3 & \cellcolor{gray!10}20.5 & \cellcolor{gray!10}26.2 & \cellcolor{gray!10}65.5 & \cellcolor{gray!10}77.8 & \cellcolor{gray!10}25.3 & \cellcolor{gray!10}2.1 & \cellcolor{gray!10}8.5 & \cellcolor{gray!10}38.6 & \cellcolor{gray!10}38.6 \\
& \cellcolor{gray!10}InternLM-7B-8k{\color{blue}*} & \cellcolor{gray!10}12.1 & \cellcolor{gray!10}16.7 & \cellcolor{gray!10}23.4 & \cellcolor{gray!10}28.7 & \cellcolor{gray!10}22.8 & \cellcolor{gray!10}9.0 & \cellcolor{gray!10}9.7 & \cellcolor{gray!10}15.9 & \cellcolor{gray!10}22.8 & \cellcolor{gray!10}52.0 & \cellcolor{gray!10}77.8 & \cellcolor{gray!10}21.2 & \cellcolor{gray!10}3.0 & \cellcolor{gray!10}6.0 & \cellcolor{gray!10}44.1 & \cellcolor{gray!10}28.8 \\
& \cellcolor{gray!10}ChatGLM2-6B-32k{\color{blue}*} & \cellcolor{gray!10}21.1 & \cellcolor{gray!10}31.5 & \cellcolor{gray!10}46.2 & \cellcolor{gray!10}45.1 & \cellcolor{gray!10}34.0 & \cellcolor{gray!10}21.9 & \cellcolor{gray!10}32.4 & \cellcolor{gray!10}24.0 & \cellcolor{gray!10}26.5 & \cellcolor{gray!10}62.5 & \cellcolor{gray!10}78.7 & \cellcolor{gray!10}36.3 & \cellcolor{gray!10}1.5 & \cellcolor{gray!10}77.0 & \cellcolor{gray!10}55.6 & \cellcolor{gray!10}49.9 \\
& \cellcolor{gray!10}ChatGLM3-6B-32k{\color{blue}*} & \cellcolor{gray!10}26.0 & \cellcolor{gray!10}43.3 & \cellcolor{gray!10}51.7 & \cellcolor{gray!10}54.4 & \cellcolor{gray!10}44.9 & \cellcolor{gray!10}40.4 & \cellcolor{gray!10}36.8 & \cellcolor{gray!10}23.9 & \cellcolor{gray!10}27.9 & \cellcolor{gray!10}79.0 & \cellcolor{gray!10}87.1 & \cellcolor{gray!10}38.2 & \cellcolor{gray!10}2.0 & \cellcolor{gray!10}99.0 & \cellcolor{gray!10}57.66 & \cellcolor{gray!10}54.76 \\
& \cellcolor{gray!10}Baichuan-13B-4k{\color{yellow}*} & \cellcolor{gray!10}0.07 & \cellcolor{gray!10}17.55 & \cellcolor{gray!10}17.28 & \cellcolor{gray!10}3.29 & \cellcolor{gray!10}15 & \cellcolor{gray!10}0.1 & \cellcolor{gray!10}6.8 & \cellcolor{gray!10}1.71 & \cellcolor{gray!10}23.1 & \cellcolor{gray!10}20.05 & \cellcolor{gray!10}20.06 & \cellcolor{gray!10}5.77 & \cellcolor{gray!10}0.06 & \cellcolor{gray!10}0.5 & \cellcolor{gray!10}47.98 & \cellcolor{gray!10}16.58 \\
& \cellcolor{gray!10}ALiBi-7B-4k{\color{yellow}*} & \cellcolor{gray!10} 0.04 & \cellcolor{gray!10} 8.13 & \cellcolor{gray!10} 17.87 & \cellcolor{gray!10} 2.73 & \cellcolor{gray!10} 8 & \cellcolor{gray!10} 1.33 & \cellcolor{gray!10} 5.31 & \cellcolor{gray!10} 1.64 & \cellcolor{gray!10} 25.55 & \cellcolor{gray!10} 9.25 & \cellcolor{gray!10} 8.83 & \cellcolor{gray!10} 4.67 & \cellcolor{gray!10} 0 & \cellcolor{gray!10} 1.27 & \cellcolor{gray!10} 46.69 & \cellcolor{gray!10} 18.54 \\
\specialrule{1pt}{2pt}{0pt}

\end{tabular}
}

\begin{tablenotes}
    \scriptsize
    \item[] \hspace{-20pt}\textsuperscript{a} Details of used LLMs in this table are presented in \cref{sec:app:models}.
\end{tablenotes}
\end{threeparttable}\vspace{-10pt}

\end{table*}

\subsection{Performance on Language Modeling Tasks}
Language modeling task is the most fundamental and the least requirement for LLMs, which is usually measured by perplexity~(PPL) on the test text data. A low PPL does not guarantee good performance on real tasks~\cite{pal2023giraffe}, however, a higher PPL suggests severe performance degradation of LLMs. We evaluate SelfExtend's language modeling performance on dataset PG19~\cite{rae2019compressive}, which contains lengthy books. PPL is used as the metric. More experimental details are presented in \cref{sec:app:setting:llt}
The results show that SelfExtend can successfully maintain a low PPL out of the pretraining context window for both Llama-2-7b-chat and Mistral. Without SelfExtend, the PPL explodes when the length of test sequence is larger than the context window. Mistral with SWA can also maintain a low PPL out of its context window. But later in the next section, we will demonstrate that a low PPL score does not necessarily indicate proficiency in handling long contexts. More discussion about PPL can be found in~\cref{apdx:ppl}. 

\begin{figure}[t]
    \centering
    \includegraphics[width=\linewidth]{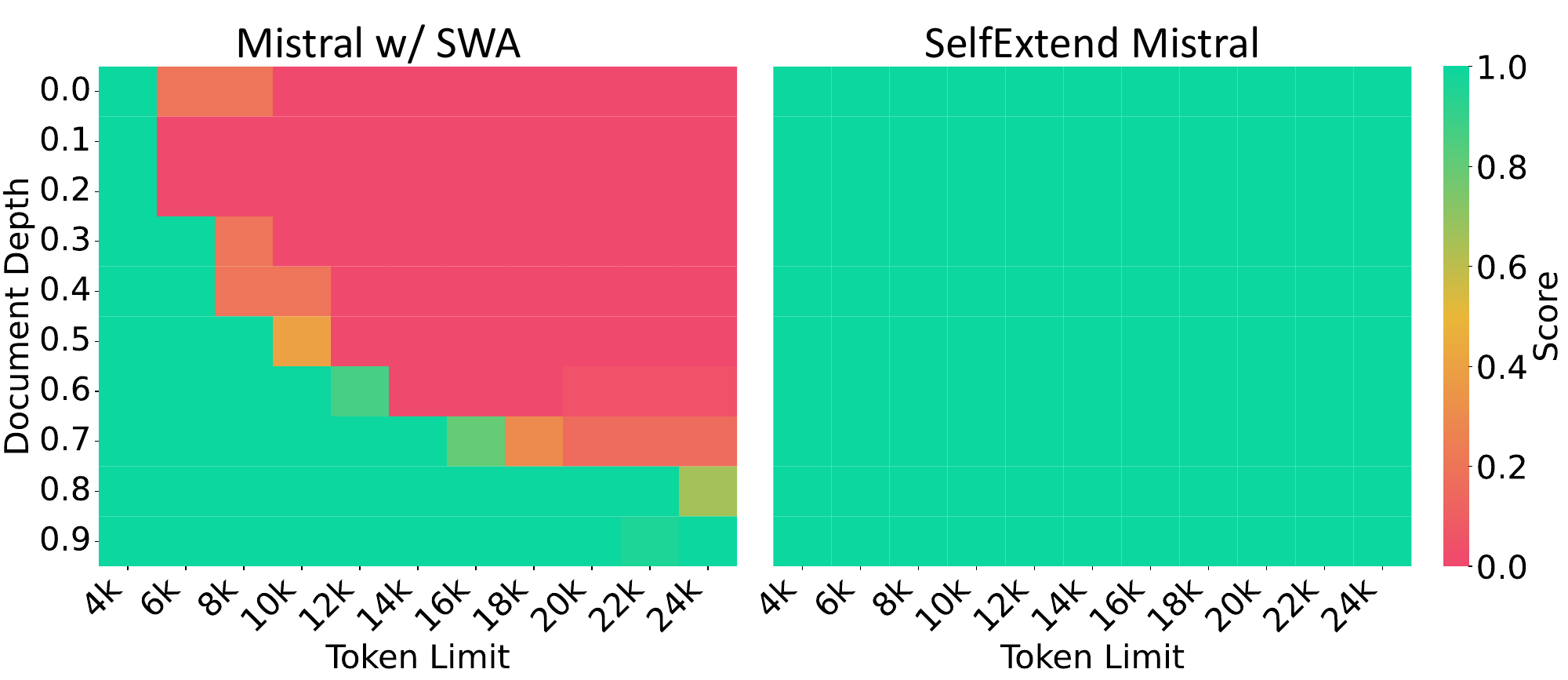}
    \vspace{-20pt}
    \caption{Passkey retrieval accuracy for Mistral-7b-instruct-0.1 with SWA or SelfExtend. Mistral with SelfExtend obtains $100\%$ passkey retrieval accuracy For all sequence length (token limit) and all depth. Mistral with SWA cannot retrieve the passkey out of the sliding window. The default sliding window size is $4096$.}
    \label{fig:passkey}
    \vspace{-5pt}
\end{figure}

\subsection{Performance on Synthetic Long Context Tasks}
The passkey retrieval task is the same as what is defined in Landmark Attention~\cite{mohtashami2023landmark}, which is a synthetic long context task. It requires a language model to retrieve a simple passkey~(i.e., a 5-digit random number) in a long meaningless text sequence. The passkey is placed with various document depths (where the passkey is placed in the input texts) and context lengths~(ranging from 4k to 24k). We tested multiple passkey retrievals for each context length and depth. The passkey was randomly placed within a span of $400$ tokens. For a depth of $0.1$ and context of 8k, the passkey was placed between tokens $800-1600$. We performed $10$ iterations per span, so $20$ total for that setting.  Experimental setting details and an example of passkey retrieval task can be found in~\cref{apdx:passkey_example}.

The results in \cref{fig:passkey} show that without any fine-tuning, SelfExtend obtains \textbf{100\%} passkey retrieval accuracy across all tested depths and context lengths. The results also demonstrate that: although Mistral w/ SWA has low PPL beyond its pretraining context window, it can only access information~(i.e. the passkey) within its sliding window. Considering the simplicity of this task, these results strongly suggest it still does not have the true ability to handle long contexts.

\begin{table}[!t]
\setlength{\tabcolsep}{2pt}
\centering
\caption{Exam evaluation results on L-Eval. \textbf{Tokens} denotes the maximum input context length. {\color{red} +} indicates the results are from us and others are reported by L-Eval. The rows in the same color~(orange, green, blue, and pink) represent the models of those rows from the same base model. The better performance between models w/ and w/o SelfExtend is highlighted in \textbf{bold}.}\label{table:l_eval}
\resizebox{1.01\linewidth}{!}{
\begin{tabular}{lccccccccc}
\toprule
\textbf{Model}  & \textbf{Tokens} & \textbf{Coursera} & \textbf{GSM} & \textbf{QuALITY} & \textbf{TOEFL}  & \textbf{CodeU} & \textbf{SFiction} &  \textbf{Avg.} \\
\midrule
Claude1.3-100k  & 100k & 60.03 & {88.00} & {73.76} & {83.64} & 17.77 & 72.65 & 65.97  \\
GPT-4-32k  & 32k & 75.58 & 96.00 & 82.17 & 84.38 & 25.55& 74.99 & 73.11 \\
Turbo-16k-0613  & 16k & 63.51 & 84.00 & 61.38 & 78.43 & 12.22 & 64.84 & 60.73 \\
\midrule
Chatglm2-6b-8k & 2k & 43.75 & 13.00 & 40.59 & 53.90& 2.22 & 54.68 & 34.69  \\
XGen-7b-8k (2k-4k-8k)  & 2k & 26.59 & 3.00 &  35.15  & 44.23 & 1.11 & 48.43 & 26.41 \\
Chatglm2-6b-8k  & 8k & 42.15 & 18.00 & 44.05 & 54.64 & 2.22 &54.68 & 35.95\\
Chatglm2-6b-32k  & 32k  &  47.81 & 27.00 & 45.04 & 55.01 & 2.22 & 57.02 & 39.01 \\
XGen-7b-8k  & 8k & 29.06 & 16.00 &  33.66  & 42.37 & 3.33 & 41.40 & 27.63\\
MPT-7b-65k & 8k & 25.23 & 8.00 & 25.24 & 17.84 & 0.00 & 39.06 & 19.22\\
\midrule
\rowcolor{blue!10}
Llama2-7b-chat & 4k  & 29.21 & 19.00 & 37.62 & 51.67 & 1.11 & 60.15 & 33.12 \\
\rowcolor{blue!10}
Longchat1.5-7b-32k & 32k  & 32.99 & 18.00 & 37.62 & 39.77 & \textbf{3.33} & 57.02 & 31.45\\
\rowcolor{blue!10}
Llama2-7b-NTK & 16k  & 32.71 & 19.00 & 33.16 & 52.78 & 0.00 & \textbf{64.84} & 33.74 \\
\rowcolor{blue!10}
SE-Llama2-7B-chat\color{red}{+}  & 16k & \textbf{35.76} &\textbf{25.00} & \textbf{41.09} & \textbf{55.39} & 1.11 & 57.81 &  \textbf{36.02}\\
\midrule
\rowcolor{blue!10}
Vicuna1.5-7b-16k  & 16k  & \textbf{38.66} & 19.00 & 39.60 & \textbf{55.39} & \textbf{5.55} & 60.15 & {36.39}  \\
\rowcolor{blue!10}
SE-Vicuna1.5-7B\color{red}{+}  & 16k & {37.21} & \textbf{21.00} & \textbf{41.58} & \textbf{55.39} & {3.33} & \textbf{63.28} & \textbf{36.96}\\
\midrule
\rowcolor{pink!10}
Llama2-13b-chat  & 4k  & 35.75 & 39.00 & \textbf{42.57} & 60.96 & 1.11 & {54.68} & {39.01} \\
\rowcolor{pink!10}
Llama2-13b-NTK & 16k  & {36.48} & 11.00 & 35.64  & 54.64&  1.11  & \textbf{63.28} & 33.69 \\
\rowcolor{pink!10}
Llama2-13b-NTK(Dyn) & 16k  & 30.08 & \textbf{43.00} & {41.58} & {64.31} & 1.11 & 35.15 & 35.87\\
\rowcolor{pink!10}
SE-Llama2-13B-chat\color{red}{+}  & 16k & \textbf{38.95} & {42.00} & 41.09 & \textbf{66.17} & 1.11 & \textbf{63.28} & \textbf{42.10} \\
\midrule
\rowcolor{green!10}
Mistral-7b-ins-0.1 w/ SWA\color{red}{+}   & 16k & 44.77 & {44.00} & 46.53 & {60.59} & 2.22 & \textbf{64.06} & {43.70}\\
\rowcolor{green!10}
Mistral-7b-ins-0.1 w/o SWA\color{red}{+}   & 8k & 43.60 & 49.00 & 45.05 & 60.59 & \textbf{4.44} & 60.94 & {43.94}\\
\rowcolor{green!10}
MistralLite\color{red}{+}  & 16k & 29.23 & 32.00 & {46.04} & 17.47 & 3.33 & 14.06 & 23.69\\
\rowcolor{green!10}
SE-Mistral-7b-ins-0.1\color{red}{+}  & 16k & \textbf{45.20} & \textbf{51.00} & \textbf{48.02} & \textbf{64.68} & 3.33 & 59.38 & \textbf{45.27} \\
\midrule

\rowcolor{cyan!10}
Phi-2\color{red}{+} & 2k & 38.37 & 64.00 & \textbf{42.08} & 55.76 & 3.33 & \textbf{52.34} & 42.64\\
\rowcolor{cyan!10}
SE-Phi-2\color{red}{+} & 8k & \textbf{42.44} & \textbf{65.00} & 41.08 & \textbf{62.83} & \textbf{4.44} & \textbf{52.34} & \textbf{44.69} \\
\midrule

\rowcolor{orange!10}
SOLAR-10.7b-Instruct-v1.0\color{red}{+} & 4k & 48.84 & \textbf{72.00} & 59.90 & 77.32 & \textbf{4.44} & 69.53 & 55.34\\
\rowcolor{orange!10}
SE-SOLAR-10.7b-v1.0\color{red}{+} & 16k &  \textbf{50.44} & \textbf{72.00} & \textbf{70.30} & \textbf{79.18} & \textbf{4.44} & \textbf{73.44} & \textbf{58.30}\\
\bottomrule
\end{tabular}
}
\vspace{-10pt}
\end{table}

\subsection{Performance on Real-World Long Context Tasks}
Evaluation solely on language modeling (measured by perplexity) and synthetic tasks like passkey retrieval cannot fully assess the long-context capabilities of LLMs. The task of Passkey retrieval is overly straightforward, and an LLM may still struggle with long context despite low perplexity. To comprehensively evaluate long-context performance, we further use two recent real-world long context benchmarks: LongBench~\cite{bai2023longbench} and  L-Eval~\cite{an2023eval}. The results are presented in \cref{tab:longbench} and \cref{table:l_eval}. On the LongBench in \cref{tab:longbench}, for all four different base LLMs and most datasets, with SelfExtend, the LLM can obtain significant performance improvements.
\\
\textbf{Llama-2-7B}: We use SelfExtend to increase Llama-2-7b-chat's context from 4k to 16k and 25k. Both significantly outperform Llama-2-7b-chat and most fine-tuned models on several datasets like HotpotQA. We also extend vicuna1.5-7B from 4k to 16k and 25k. With SelfExtend, vicuna1.5-7B surpasses its fine-tuned counterpart vicuna1.5-7B-16k and ranks among top Llama-2-7b models.
On some datasets, the 25k variant underperforms the 16k one due to the trade-off between larger context and positional precision. More details about the trade-off is in \cref{sec:ablation}.
\\
\textbf{Mistral-7B}: We extend Mistral-7B's context to 16k, significantly improving its long context ability over the base model. The fine-tuned variant MistralLite (\cite{amazon2023mistrallite}) achieves the best performance on most datasets. However, many of these datasets were included in MistralLite's fine-tuning data, such as NarrativeQA\footnote{More details about MistralLite's fine-tuning data can be found at \url{https://huggingface.co/amazon/MistralLite}. At least, GovReport, QMSum, NarrativeQA, Qasper, QuALITY, and HotpotQA are included. Meanwhile, Multi-passage QA and summarization tasks are also in fine-tuning data. This also violates zero-shot evaluation conditions.}.
\\
\textbf{SOLAR-10.7B and Phi-2}: They have no finetuned variant for context window extension yet. SelfExtend can also obtain substantial performance improvements.

On the LEval benchmark in \cref{table:l_eval}, we observe similar results. Compared to fine-tuning free baselines like NTK or further fine-tuned models like Longchat1.5-7b-32k and Vicuna1.5-7b-32k, SelfExtend achieves superior performance on nearly all datasets\footnote{LEval performance seems sensitive to prompt engineering for these sub-13B LLMs. For example, on some datasets, vanilla vicuna-13b underperforms vanilla vicuna-7b.}.

In summary, on the two benchmarks, \textbf{SelfExtend achieves comparable or better performance, compared to methods that requires further fine-tuning}. Despite our initial expectation being that SelfExtend would simply outperform the base model without additional extension methods, it is remarkable that our SelfExtend, which solely operates during inference without the need for fine-tuning or training, achieves such impressive performance.

\subsection{Performance on Short Context Tasks}
We argue that an ideal context length extension method should not degrade performance on standard short-context tasks. Previous fine-tuning based methods usually undergo performance degradation on short-context tasks~\cite{peng2023yarn, xiong2023effective}. Following~\cite{peng2023yarn}, we use Hugging Face Open LLM Leaderboard~\cite{eval-harness} to evaluate SelfExtend's performance on five public short context tasks. Specifically, we use 25-shot ARC-Challenge~\cite{clark2018think}, 10-shot
HellaSwag~\cite{zellers2019hellaswag}, 5-shot MMLU~\cite{hendrycks2020measuring}, 0-shot TruthfulQA~\cite{lin2021truthfulqa}, and 5-shot GSM8K~\cite{cobbe2021gsm8k}. The results are shown in~\cref{tab:open_llm}. We also investigate the influence of varying group sizes and neighbor window sizes on short-context tasks and we present the results in~\cref{apdx:benchmark}. 

The results show that SelfExtend can maintain the performance of the short-context tasks, while enhance the performance on long-context tasks.  Moreover, because SeldExtend does not require any fine-tuning and only takes effect during inference, SelfExtend can be readily adopted as a plug-in component for LLMs. This means SelfExtend can be automatically and inherently disabled while encountering short-text sequences. Then, with the parameters remaining unchanged, LLMs can maintain its original inference mechanism on those short-context scenarios.

\begin{table}[!t]
    \fontsize{15}{20}\selectfont
    \centering
    \vspace{-15pt}
    \caption{Performance of SelfExtend on Hugging Face Open LLM benchmark compared to baselines: Llama 2, Llama-2-chat-4, Mistral-instruct-v0.1 and Phi-2. We use the same hyper-parameters as on LongBench benchmark. For Llama-2 \& Llama-2-chat based SelfExtend, the group size is $16$ and neighbor window is $1024$; for Mistral based SelfExtend, the group size is $6$ and neighbor window is $1024$; for Phi-2 based SelfExtend, the group size is $12$ and neighbor window is $512$.}\label{tab:open_llm}
    \resizebox{\linewidth}{!}{
    \begin{tabular}{llrccccccc}
       \toprule
       Size &  Name  & ARC-c & Hellaswag & MMLU & TruthfulQA & GSM8k  \\
       \midrule
        7B &       Llama-2  & \textbf{52.99} & \textbf{78.66} & 46.58 & \textbf{38.97} & \textbf{14.94} \\
        7B &       SE-Llama 2 & \textbf{52.99} & 78.65& \textbf{46.68} & \textbf{38.97} & 14.71 \\
       \midrule 
        7B &       Llama-2-chat & \textbf{52.73} & \textbf{78.49} & \textbf{48.20}  & 45.32 & 18.73\\
        7B &       SE-Llama-2-chat-16k & \textbf{52.73} & \textbf{78.49} & 48.09 & \textbf{45.33} & \textbf{18.88} \\
       \midrule 
        7B &      Mistral-instruct-v0.1 & 54.35 & \textbf{75.72} & 55.57 & \textbf{55.89} & 30.93 \\
        7B &      SE-Mistral-instruct-v0.1 & \textbf{54.44} & 75.71 & \textbf{55.59} & \textbf{55.89} & \textbf{31.39} \\
       \midrule 
        2.7B &    Phi-2 & \textbf{61.17} & 75.13 & 58.20 & \textbf{44.54} & 55.11 \\
        2.7B &    SE-Phi-2 & 61.00 & \textbf{75.20} & \textbf{58.29} & \textbf{44.54} & \textbf{55.42} \\
       \bottomrule
    \end{tabular}
    } \vspace{-22pt}
\end{table}

\subsection{Ablations on Group Size and Neighbor Window}\label{sec:ablation}
We investigate the influence of varying the group size~$G_s$ and the neighbor window~$w_n$. We experiments with Phi-2 on four real-world datasets from Longbench: narrativeqa, qasper, triviaqa, and repobench-p. The results are presented in~\cref{fig:hype}. Form the results, we observe two trade-offs:

\begin{figure*}[!t]
    \centering
    \vspace{-10pt}
    \includegraphics[width=\textwidth]{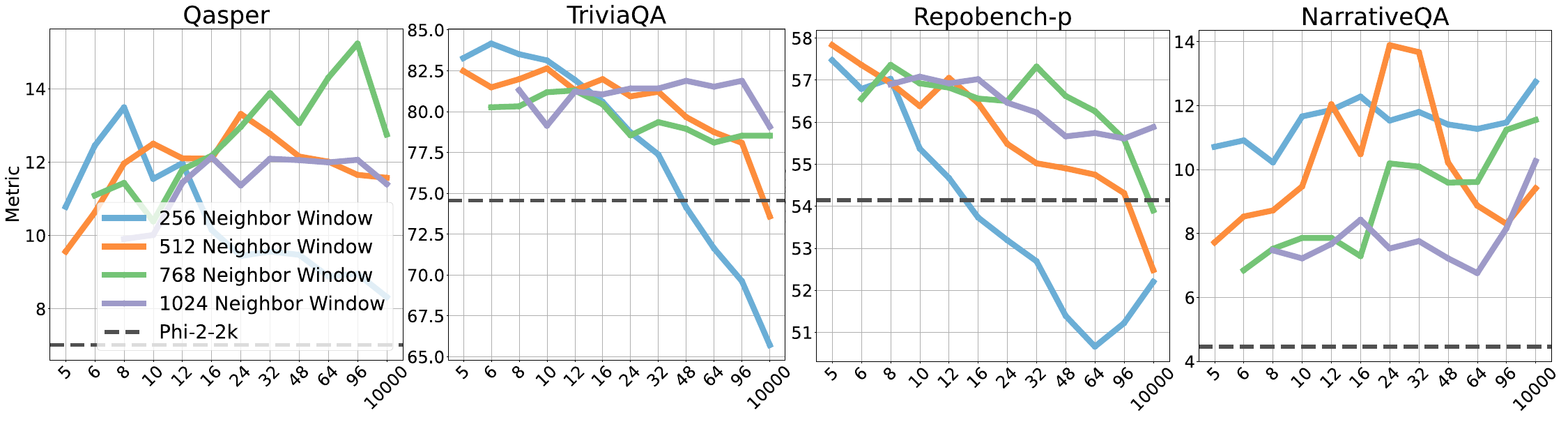}
    \vspace{-28pt}
    \caption{The performance of Phi-2 when utilizing SelfExtend to extend its context window length to 8k, with varying group sizes and neighbor window sizes.  The y-axis indicates performance and the x-axis shows the group size. And neighbor window size is from $256, 512, 768, 1024$. Group size of $10000$ in this experiment means all tokens out of the neighbor window are in the same group~(10000$>$8k). Some combination~(e.g. $G_s=6$ \& $w_n=1024$) is omitted if the corresponding extended context window~(\cref{eq:length}) is smaller than 8k. The dashed line is the performance of vanilla phi-2 with a 2k context window size.}
    \label{fig:hype}
    \vspace{-15pt}
\end{figure*}

1) There is a trade-off with respect to group size in SelfExtend. Generally, both too small and too large group sizes can result in inferior performance compared to an optimal level. With a large group size, position information becomes more coarse, potentially causing performance drops. Conversely, small group sizes require SelfExtend to utilize larger position embeddings to extend the context window. These larger position embeddings are less trained compared to smaller ones. For example, in Llama-2 with its 4096 context window, the relative position 4095 accounts for only 1/2048 the frequency of the relative  position 2048 in training. These under-trained relative positions can also degrade performance. This trade-off produces the 'peak' shape in the figure, indicating the extended context window differs from the ideal case described in \cref{eq:length}.

2) There is also another trade-off w.r.t. neighbor window size. With larger neighbor window sizes, there is more precise information about neighbor tokens, which is the most important. But a larger neighbor window size means SelfExtend has to use a larger group size for a long sequence, compared to using a smaller neighbor window size \& smaller group size, the information about the whole sequence becomes coarse. 

More results can be found in \cref{apdx:hypara}. When the group size is so large that all distant tokens are in one group, SelfExtend degenerates into ReRoPE~(e.g. group size of 32768 in \cref{fig:hypara}).

\begin{table}[!t]
    \fontsize{15}{20}\selectfont
    \vspace{-15pt}
    \centering
    \caption{Performance of Phi-2 with different context window lengths. The vanilla Phi-2 has a 2k context window. SelfExtend extends Phi-2 to 4k~($G_s=4$,$w_n=512$), 6k~($G_s=8$,$w_n=512$) and 8k~($G_s=12$,$w_n=512$). The performance improvement compared to vanilla Phi-2 is in the parenthesis.}
    \label{tab:length}
    \resizebox{\linewidth}{!}{
    \label{tab:detailed_performance_comparison}
    \begin{tabular}{l|cccc}
    \hline
    Context Length & 2k~(vanilla) & 4k & 6k & 8k \\ 
    \hline
    \multicolumn{5}{c}{Document QA} \\
    \hline
    NarrativeQA & 4.46 & 6.49 (+45.52\%) & 8.98 (+101.35\%) & 12.04 (+169.96\%) \\
    Qasper & 7.01 & 11.16 (+59.20\%) & 12.84 (+83.17\%) & 12.10 (+72.61\%) \\
    \hline
    \multicolumn{5}{c}{Summarization} \\
    \hline
    Gov\_report & 25.46 & 27.91 (+9.62\%) & 28.14 (+10.53\%) & 27.51 (+8.05\%) \\
    Qmsum & 14.32 & 14.88 (+3.91\%) & 16.72 (+16.76\%) & 18.58 (+29.75\%) \\
    \hline
    \multicolumn{5}{c}{Few-shot Learning} \\
    \hline
    Trec & 50.5 & 60.0 (+18.81\%) & 62.5 (+23.76\%) & 60.0 (+18.81\%) \\
    Triviaqa & 74.55 & 84.88 (+13.86\%) & 82.64 (+10.85\%) & 81.31 (+9.07\%) \\
    \hline
    \multicolumn{5}{c}{Coding} \\
    \hline
    Repobench-p & 54.14 & 56.18 (+3.77\%) & 56.76 (+4.84\%) & 57.05 (+5.37\%) \\
    Lcc & 58.96 & 59.06 (+0.17\%) & 58.88 (-0.14\%) & 59.42 (+0.78\%) \\
    \hline
    \end{tabular}
    }
    \vspace{-20pt}
\end{table}

\subsection{Performance with Varying Context Window Length}
To validate SelfExtend's efficacy in enabling LLMs to utilize extended context windows, we assess Phi-2's performance across varying context lengths with SelfExtend, referencing \cref{tab:length}. Across four task types from LongBench, results are generally improved with longer contexts. Notably, SelfExtend monotonically enhances performance on NarrativeQA and Qmsum. While significant improvements are observed across most datasets, a 'peak' in performance suggests a trade-off, as discussed in \cref{sec:ablation}: longer contexts offer more relevant information, but the larger group sizes required by SelfExtend to extend the context window may cause less precise positional information\footnote{Other possible reasons include: Phi-2 is a base model without instruction tuning, and SelfExtend's performance is not  optimal as we use the same set of hyperparameters across all datasets, which cannot showcase SelfExtend's full potential}. Regarding Lcc, performance remains consistent, possibly due to its reliance on local codes and shorter dataset lengths\footnote{With Phi-2 tokenizer, over $60\%$ of Lcc instances are under 4096 tokens, with an average length of 4069.7}.

\begin{figure}[t]
    \centering
    \vspace{-5pt}
    \includegraphics[width=0.9\linewidth]{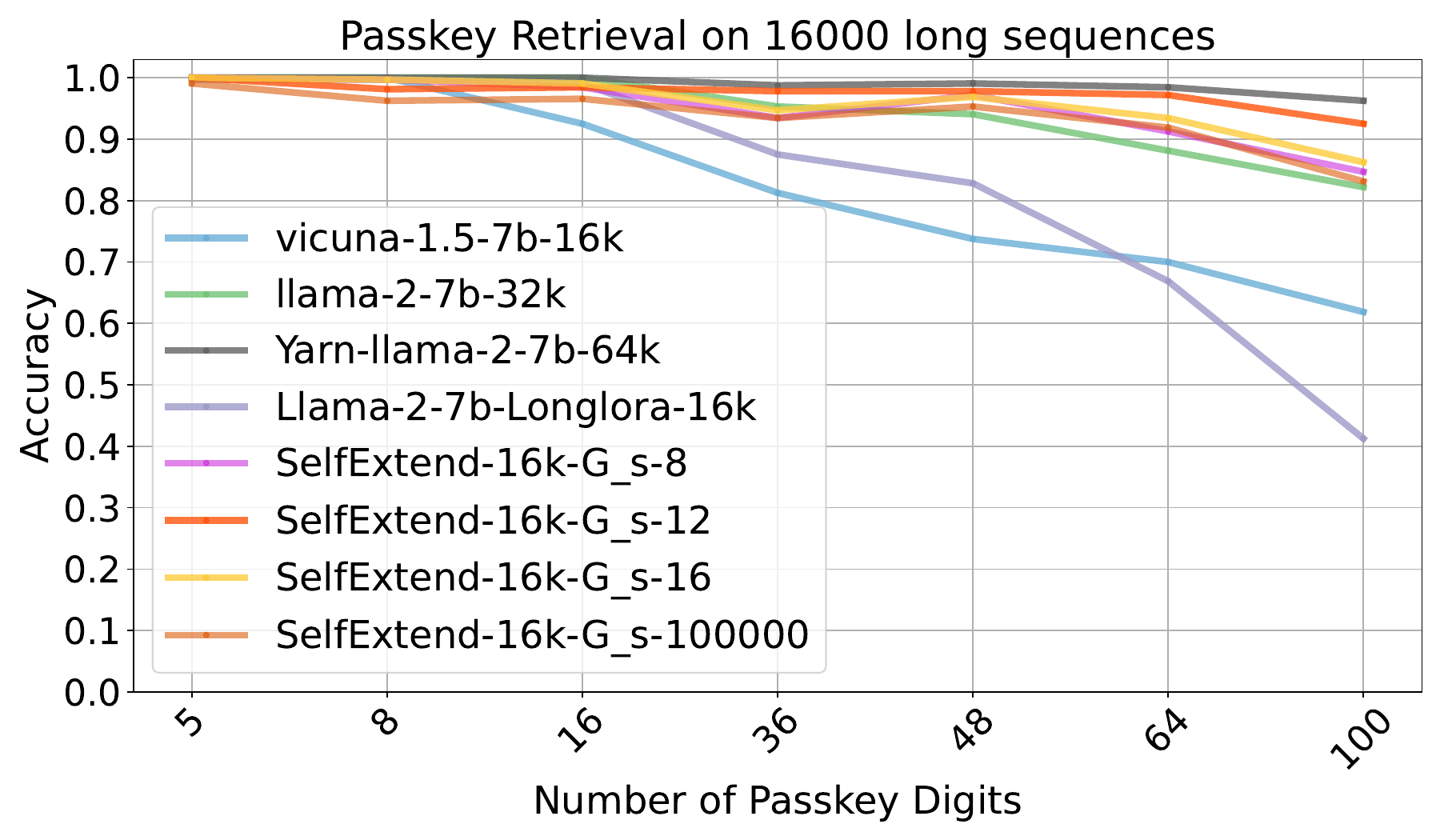}
    \vspace{-15pt}
    \caption{Passkey retrieval accuracy for four fine-tuning-based long-context models and SelfExtend on Llama-2-chat-7b across four group sizes: $8, 12, 16,$ and $100000$. For SelfExtend, the neighbor window is $1024$. A group size of $100000$ indicates that all tokens outside the neighbor window are in the same group.}
    \label{fig:long_passkey}
    \vspace{-20pt}
\end{figure}

\subsection{Varying-Length Passkey Retrieval Task}
The conventional passkey retrieval task, along with prevalent benchmark datasets, primarily assesses the proficiency of LLMs in identifying and leveraging pertinent information. 
Traditionally, this task involves passkeys not exceeding 5 digits in length. To evaluate the LLMs' capabilities of producing consistent and precise outcomes for long sequences, we extended the task to incorporate passkeys with larger lengths. We test passkeys in $5, 8, 16, 36, 48, 64, 100$ digits. The input sequence contains $16,000$ characters. More details are presented in \cref{sec:app:setting:lprt}.

The results, depicted in \cref{fig:long_passkey}, illustrate a common trend: while short passkeys of 5 or 8 digits are easily managed by all, divergences in performance emerge as the length of passkey increases. Notably, with the exception of Yarn, many tuning-based methods are unable to accurately reproduce passkeys beyond 64 digits, and some of them even experience a marked decline in performance when the passkey length exceeds 16 digits. Remarkably, although without tuning, SelfExtend maintains its superiority.  These findings suggest that we should carefully choose the training approach when fine-tuning models to handle long contexts.

\section{Conclusion and Discussion}

In this paper, we argue that LLMs themselves have the inherent ability to handle long sequences and propose SelfExtend to elicit the inherent long context abilities for LLMs by mapping unseen relative positions into those seen during pretraining. Without any tuning or further training, SelfExtend can effectively improve LLMs' long context performance.

\emph{Limitations:} SelfExtend increases computation cost with naive implementations, since it performs extra attention across all query-key pairs. However, with optimizations like blocked kernels~(e.g. Flash Attention \cite{dao2022flashattention}), this becomes linear rather than quadratic, and the marginal cost is small enough to be ignored for long input sequences. Also, the performance degrades with large group size, preventing indefinitely long contexts.
Besides, SelfExtend still processes the entire sequence to ensure information integrity, while some methods such as prompt compression~\cite{chuang2024learning, jiang-etal-2023-longllmlingua} can shorten the input to reduce computation.
Additionally, evaluation methodologies for assessing long context abilities remain open research questions. Standard practices have yet to emerge, complicating experimental results.

\emph{Future Work:} We are interested in more sophisticated mapping methods to replace the simple \textsc{floor} operation, aiming to enhance long context understanding and extend the context window length. Additionally, we plan to further investigate the complex behaviors of LLMs using SelfExtend.

\section*{Impact Statement}
The impact of this contribution is multifaceted. Firstly, it enhances the usability and applicability of LLMs across various domains that require the processing of lengthy input sequences, such as document analysis, long-form question answering, and retrieval augmented generation. Secondly, the ability to extend the context window without fine-tuning simplifies the deployment of advanced language models, making them more accessible to a broader range of users and applications. The availability of the code makes this solution readily accessible to researchers and practitioners, promising widespread adoption and further innovation in the field of natural language processing.

\section*{Acknowledgements}
The authors thank the anonymous reviewers for their helpful comments. This work is in part supported by NSF grants NSF IIS-2310260 and IIS-2224843. 
The views and conclusions contained in this paper are those of the authors and should not be interpreted as representing any funding agencies.

\bibliography{main.bib}

\begin{thebibliography}{53}
\providecommand{\natexlab}[1]{#1}
\providecommand{\url}[1]{\texttt{#1}}
\expandafter\ifx\csname urlstyle\endcsname\relax
  \providecommand{\doi}[1]{doi: #1}\else
  \providecommand{\doi}{doi: \begingroup \urlstyle{rm}\Url}\fi

\bibitem[amazon(2023)]{amazon2023mistrallite}
amazon.
\newblock Mistrallite model.
\newblock \url{https://huggingface.co/amazon/MistralLite}, 2023.
\newblock [Online; accessed 29-December-2023].

\bibitem[An et~al.(2023)An, Gong, Zhong, Li, Zhang, Kong, and Qiu]{an2023eval}
An, C., Gong, S., Zhong, M., Li, M., Zhang, J., Kong, L., and Qiu, X.
\newblock L-eval: Instituting standardized evaluation for long context language models.
\newblock \emph{arXiv preprint arXiv:2307.11088}, 2023.

\bibitem[Anothropic(2023)]{needle_claudes}
Anothropic.
\newblock Long context prompting for claude 2.1.
\newblock \url{https://www.anthropic.com/news/claude-2-1-prompting}, 2023.

\bibitem[Bai et~al.(2021)Bai, Luo, Zhao, Wen, and Wang]{bai2021recent}
Bai, T., Luo, J., Zhao, J., Wen, B., and Wang, Q.
\newblock Recent advances in adversarial training for adversarial robustness.
\newblock \emph{arXiv preprint arXiv:2102.01356}, 2021.

\bibitem[Bai et~al.(2023)Bai, Lv, Zhang, Lyu, Tang, Huang, Du, Liu, Zeng, Hou, et~al.]{bai2023longbench}
Bai, Y., Lv, X., Zhang, J., Lyu, H., Tang, J., Huang, Z., Du, Z., Liu, X., Zeng, A., Hou, L., et~al.
\newblock Longbench: A bilingual, multitask benchmark for long context understanding.
\newblock \emph{arXiv preprint arXiv:2308.14508}, 2023.

\bibitem[Brown et~al.(2020)Brown, Mann, Ryder, Subbiah, Kaplan, Dhariwal, Neelakantan, Shyam, Sastry, Askell, et~al.]{brown2020language}
Brown, T., Mann, B., Ryder, N., Subbiah, M., Kaplan, J.~D., Dhariwal, P., Neelakantan, A., Shyam, P., Sastry, G., Askell, A., et~al.
\newblock Language models are few-shot learners.
\newblock \emph{Advances in neural information processing systems}, 33:\penalty0 1877--1901, 2020.

\bibitem[Chen et~al.(2023{\natexlab{a}})Chen, Li, Meng, Liang, and Bing]{chen2023clex}
Chen, G., Li, X., Meng, Z., Liang, S., and Bing, L.
\newblock Clex: Continuous length extrapolation for large language models.
\newblock \emph{arXiv preprint arXiv:2310.16450}, 2023{\natexlab{a}}.

\bibitem[Chen et~al.(2023{\natexlab{b}})Chen, Wong, Chen, and Tian]{chen2023extending}
Chen, S., Wong, S., Chen, L., and Tian, Y.
\newblock Extending context window of large language models via positional interpolation.
\newblock \emph{arXiv preprint arXiv:2306.15595}, 2023{\natexlab{b}}.

\bibitem[Chen et~al.(2023{\natexlab{c}})Chen, Qian, Tang, Lai, Liu, Han, and Jia]{chen2023longlora}
Chen, Y., Qian, S., Tang, H., Lai, X., Liu, Z., Han, S., and Jia, J.
\newblock Longlora: Efficient fine-tuning of long-context large language models.
\newblock \emph{arXiv preprint arXiv:2309.12307}, 2023{\natexlab{c}}.

\bibitem[Chuang et~al.(2024)Chuang, Xing, Chang, Liu, Chen, and Hu]{chuang2024learning}
Chuang, Y.-N., Xing, T., Chang, C.-Y., Liu, Z., Chen, X., and Hu, X.
\newblock Learning to compress prompt in natural language formats.
\newblock \emph{arXiv preprint arXiv:2402.18700}, 2024.

\bibitem[Clark et~al.(2018)Clark, Cowhey, Etzioni, Khot, Sabharwal, Schoenick, and Tafjord]{clark2018think}
Clark, P., Cowhey, I., Etzioni, O., Khot, T., Sabharwal, A., Schoenick, C., and Tafjord, O.
\newblock Think you have solved question answering? try arc, the ai2 reasoning challenge.
\newblock \emph{arXiv preprint arXiv:1803.05457}, 2018.

\bibitem[Cobbe et~al.(2021)Cobbe, Kosaraju, Bavarian, Chen, Jun, Kaiser, Plappert, Tworek, Hilton, Nakano, Hesse, and Schulman]{cobbe2021gsm8k}
Cobbe, K., Kosaraju, V., Bavarian, M., Chen, M., Jun, H., Kaiser, L., Plappert, M., Tworek, J., Hilton, J., Nakano, R., Hesse, C., and Schulman, J.
\newblock Training verifiers to solve math word problems.
\newblock \emph{arXiv preprint arXiv:2110.14168}, 2021.

\bibitem[Dai et~al.(2019)Dai, Yang, Yang, Carbonell, Le, and Salakhutdinov]{dai2019transformer}
Dai, Z., Yang, Z., Yang, Y., Carbonell, J., Le, Q.~V., and Salakhutdinov, R.
\newblock Transformer-xl: Attentive language models beyond a fixed-length context.
\newblock \emph{arXiv preprint arXiv:1901.02860}, 2019.

\bibitem[Dao et~al.(2022)Dao, Fu, Ermon, Rudra, and R{\'e}]{dao2022flashattention}
Dao, T., Fu, D., Ermon, S., Rudra, A., and R{\'e}, C.
\newblock Flashattention: Fast and memory-efficient exact attention with io-awareness.
\newblock \emph{Advances in Neural Information Processing Systems}, 35:\penalty0 16344--16359, 2022.

\bibitem[Gao et~al.(2023)Gao, Tow, Abbasi, Biderman, Black, DiPofi, Foster, Golding, Hsu, Le~Noac'h, Li, McDonell, Muennighoff, Ociepa, Phang, Reynolds, Schoelkopf, Skowron, Sutawika, Tang, Thite, Wang, Wang, and Zou]{eval-harness}
Gao, L., Tow, J., Abbasi, B., Biderman, S., Black, S., DiPofi, A., Foster, C., Golding, L., Hsu, J., Le~Noac'h, A., Li, H., McDonell, K., Muennighoff, N., Ociepa, C., Phang, J., Reynolds, L., Schoelkopf, H., Skowron, A., Sutawika, L., Tang, E., Thite, A., Wang, B., Wang, K., and Zou, A.
\newblock A framework for few-shot language model evaluation, 12 2023.
\newblock URL \url{https://zenodo.org/records/10256836}.

\bibitem[gkamradt(2023)]{gkamradt2023needle}
gkamradt.
\newblock Llmtest\_needleinahaystack: Doing simple retrieval from llm models.
\newblock \url{https://github.com/gkamradt/LLMTest_NeedleInAHaystack/tree/main}, 2023.
\newblock [Online; accessed 29-December-2023].

\bibitem[Han et~al.(2023)Han, Wang, Xiong, Chen, Ji, and Wang]{han2023lm}
Han, C., Wang, Q., Xiong, W., Chen, Y., Ji, H., and Wang, S.
\newblock Lm-infinite: Simple on-the-fly length generalization for large language models.
\newblock \emph{arXiv preprint arXiv:2308.16137}, 2023.

\bibitem[Hendrycks et~al.(2020)Hendrycks, Burns, Basart, Zou, Mazeika, Song, and Steinhardt]{hendrycks2020measuring}
Hendrycks, D., Burns, C., Basart, S., Zou, A., Mazeika, M., Song, D., and Steinhardt, J.
\newblock Measuring massive multitask language understanding.
\newblock \emph{arXiv preprint arXiv:2009.03300}, 2020.

\bibitem[Javaheripi et~al.(2023)Javaheripi, Bubeck, Abdin, Aneja, Bubeck, Mendes, Chen, Giorno, Eldan, Gopi, Gunasekar, Javaheripi, Kauffmann, Lee, Li, Nguyen, de~Rosa, Saarikivi, Salim, Shah, Santacroce, Behl, Kalai, Wang, Ward, Witte, Zhang, and Zhang]{javaheripi2023phi}
Javaheripi, M., Bubeck, S., Abdin, M., Aneja, J., Bubeck, S., Mendes, C. C.~T., Chen, W., Giorno, A.~D., Eldan, R., Gopi, S., Gunasekar, S., Javaheripi, M., Kauffmann, P., Lee, Y.~T., Li, Y., Nguyen, A., de~Rosa, G., Saarikivi, O., Salim, A., Shah, S., Santacroce, M., Behl, H.~S., Kalai, A.~T., Wang, X., Ward, R., Witte, P., Zhang, C., and Zhang, Y.
\newblock Phi-2: The surprising power of small language models, 2023.

\bibitem[Jiang et~al.(2023{\natexlab{a}})Jiang, Sablayrolles, Mensch, Bamford, Chaplot, Casas, Bressand, Lengyel, Lample, Saulnier, et~al.]{jiang2023mistral}
Jiang, A.~Q., Sablayrolles, A., Mensch, A., Bamford, C., Chaplot, D.~S., Casas, D. d.~l., Bressand, F., Lengyel, G., Lample, G., Saulnier, L., et~al.
\newblock Mistral 7b.
\newblock \emph{arXiv preprint arXiv:2310.06825}, 2023{\natexlab{a}}.

\bibitem[Jiang et~al.(2023{\natexlab{b}})Jiang, Wu, , Luo, Li, Lin, Yang, and Qiu]{jiang-etal-2023-longllmlingua}
Jiang, H., Wu, Q., , Luo, X., Li, D., Lin, C.-Y., Yang, Y., and Qiu, L.
\newblock {L}ong{LLML}ingua: Accelerating and enhancing llms in long context scenarios via prompt compression.
\newblock \emph{ArXiv preprint}, abs/2310.06839, 2023{\natexlab{b}}.
\newblock URL \url{https://arxiv.org/abs/2310.06839}.

\bibitem[Ke et~al.(2020)Ke, He, and Liu]{ke2020rethinking}
Ke, G., He, D., and Liu, T.-Y.
\newblock Rethinking positional encoding in language pre-training.
\newblock \emph{arXiv preprint arXiv:2006.15595}, 2020.

\bibitem[Kim et~al.(2023)Kim, Park, Kim, Lee, Song, Kim, Kim, Kim, Lee, Kim, et~al.]{kim2023solar}
Kim, D., Park, C., Kim, S., Lee, W., Song, W., Kim, Y., Kim, H., Kim, Y., Lee, H., Kim, J., et~al.
\newblock Solar 10.7 b: Scaling large language models with simple yet effective depth up-scaling.
\newblock \emph{arXiv preprint arXiv:2312.15166}, 2023.

\bibitem[Lin et~al.(2021)Lin, Hilton, and Evans]{lin2021truthfulqa}
Lin, S., Hilton, J., and Evans, O.
\newblock Truthfulqa: Measuring how models mimic human falsehoods.
\newblock \emph{arXiv preprint arXiv:2109.07958}, 2021.

\bibitem[Liu et~al.(2021)Liu, Shen, He, Zhang, Xu, Yu, and Cui]{liu2021towards}
Liu, J., Shen, Z., He, Y., Zhang, X., Xu, R., Yu, H., and Cui, P.
\newblock Towards out-of-distribution generalization: A survey.
\newblock \emph{arXiv preprint arXiv:2108.13624}, 2021.

\bibitem[Liu et~al.(2024)Liu, Yuan, Jin, Zhong, Xu, Braverman, Chen, and Hu]{liu2024kivi}
Liu, Z., Yuan, J., Jin, H., Zhong, S., Xu, Z., Braverman, V., Chen, B., and Hu, X.
\newblock Kivi: A tuning-free asymmetric 2bit quantization for kv cache.
\newblock \emph{arXiv preprint arXiv:2402.02750}, 2024.

\bibitem[Mohtashami \& Jaggi(2023)Mohtashami and Jaggi]{mohtashami2023landmark}
Mohtashami, A. and Jaggi, M.
\newblock Landmark attention: Random-access infinite context length for transformers.
\newblock \emph{arXiv preprint arXiv:2305.16300}, 2023.

\bibitem[Pal et~al.(2023)Pal, Karkhanis, Roberts, Dooley, Sundararajan, and Naidu]{pal2023giraffe}
Pal, A., Karkhanis, D., Roberts, M., Dooley, S., Sundararajan, A., and Naidu, S.
\newblock Giraffe: Adventures in expanding context lengths in llms.
\newblock \emph{arXiv preprint arXiv:2308.10882}, 2023.

\bibitem[Peng et~al.(2023)Peng, Quesnelle, Fan, and Shippole]{peng2023yarn}
Peng, B., Quesnelle, J., Fan, H., and Shippole, E.
\newblock Yarn: Efficient context window extension of large language models.
\newblock \emph{arXiv preprint arXiv:2309.00071}, 2023.

\bibitem[Press et~al.(2021)Press, Smith, and Lewis]{press2021train}
Press, O., Smith, N.~A., and Lewis, M.
\newblock Train short, test long: Attention with linear biases enables input length extrapolation.
\newblock \emph{arXiv preprint arXiv:2108.12409}, 2021.

\bibitem[Rae et~al.(2019)Rae, Potapenko, Jayakumar, and Lillicrap]{rae2019compressive}
Rae, J.~W., Potapenko, A., Jayakumar, S.~M., and Lillicrap, T.~P.
\newblock Compressive transformers for long-range sequence modelling.
\newblock \emph{arXiv preprint arXiv:1911.05507}, 2019.

\bibitem[Raffel et~al.(2020)Raffel, Shazeer, Roberts, Lee, Narang, Matena, Zhou, Li, and Liu]{raffel2020exploring}
Raffel, C., Shazeer, N., Roberts, A., Lee, K., Narang, S., Matena, M., Zhou, Y., Li, W., and Liu, P.~J.
\newblock Exploring the limits of transfer learning with a unified text-to-text transformer.
\newblock \emph{The Journal of Machine Learning Research}, 21\penalty0 (1):\penalty0 5485--5551, 2020.

\bibitem[Rozi{\`e}re et~al.(2023)Rozi{\`e}re, Gehring, Gloeckle, Sootla, Gat, Tan, Adi, Liu, Remez, Rapin, et~al.]{roziere2023code}
Rozi{\`e}re, B., Gehring, J., Gloeckle, F., Sootla, S., Gat, I., Tan, X.~E., Adi, Y., Liu, J., Remez, T., Rapin, J., et~al.
\newblock Code llama: Open foundation models for code.
\newblock \emph{arXiv preprint arXiv:2308.12950}, 2023.

\bibitem[Shen et~al.(2021)Shen, Liu, He, Zhang, Xu, Yu, and Cui]{shen2021towards}
Shen, Z., Liu, J., He, Y., Zhang, X., Xu, R., Yu, H., and Cui, P.
\newblock Towards out-of-distribution generalization: A survey.
\newblock \emph{arXiv preprint arXiv:2108.13624}, 2021.

\bibitem[Shi et~al.(2021)Shi, Gao, Ren, Xu, Liang, Li, and Kwok]{shi2021sparsebert}
Shi, H., Gao, J., Ren, X., Xu, H., Liang, X., Li, Z., and Kwok, J. T.-Y.
\newblock Sparsebert: Rethinking the importance analysis in self-attention.
\newblock In \emph{International Conference on Machine Learning}, pp.\  9547--9557. PMLR, 2021.

\bibitem[Su(2023)]{rerope2023}
Su, J.
\newblock Rectified rotary position embeddings.
\newblock \url{https://github.com/bojone/rerope}, 2023.

\bibitem[Su et~al.(2022)Su, Lu, Pan, Murtadha, Wen, and Liu]{su2022roformer}
Su, J., Lu, Y., Pan, S., Murtadha, A., Wen, B., and Liu, Y.
\newblock Ro{F}ormer: Enhanced transformer with rotary position embedding, 2022.
\newblock arXiv: 2104.09864.

\bibitem[Sun et~al.(2023)Sun, Dong, Patra, Ma, Huang, Benhaim, Chaudhary, Song, and Wei]{xpos}
Sun, Y., Dong, L., Patra, B., Ma, S., Huang, S., Benhaim, A., Chaudhary, V., Song, X., and Wei, F.
\newblock A length-extrapolatable transformer.
\newblock In Rogers, A., Boyd-Graber, J., and Okazaki, N. (eds.), \emph{Proceedings of the 61st Annual Meeting of the Association for Computational Linguistics (Volume 1: Long Papers)}, pp.\  14590--14604, Toronto, Canada, July 2023. Association for Computational Linguistics.
\newblock \doi{10.18653/v1/2023.acl-long.816}.
\newblock URL \url{https://aclanthology.org/2023.acl-long.816}.

\bibitem[Team(2023)]{MosaicML2023Introducing}
Team, M.~N.
\newblock Introducing mpt-7b: A new standard for open-source, commercially usable llms, 2023.
\newblock URL \url{www.mosaicml.com/blog/mpt-7b}.
\newblock Accessed: 2023-05-05.

\bibitem[Touvron et~al.(2023)Touvron, Martin, Stone, Albert, Almahairi, Babaei, Bashlykov, Batra, Bhargava, Bhosale, et~al.]{touvron2023llama}
Touvron, H., Martin, L., Stone, K., Albert, P., Almahairi, A., Babaei, Y., Bashlykov, N., Batra, S., Bhargava, P., Bhosale, S., et~al.
\newblock Llama 2: Open foundation and fine-tuned chat models.
\newblock \emph{arXiv preprint arXiv:2307.09288}, 2023.

\bibitem[Vaswani et~al.(2017)Vaswani, Shazeer, Parmar, Uszkoreit, Jones, Gomez, Kaiser, and Polosukhin]{vaswani2017attention}
Vaswani, A., Shazeer, N., Parmar, N., Uszkoreit, J., Jones, L., Gomez, A.~N., Kaiser, {\L}., and Polosukhin, I.
\newblock Attention is all you need.
\newblock \emph{Advances in neural information processing systems}, 30, 2017.

\bibitem[Wu et~al.(2021)Wu, Peng, Chen, Fu, and Chao]{wu2021rethinking}
Wu, K., Peng, H., Chen, M., Fu, J., and Chao, H.
\newblock Rethinking and improving relative position encoding for vision transformer.
\newblock In \emph{Proceedings of the IEEE/CVF International Conference on Computer Vision}, pp.\  10033--10041, 2021.

\bibitem[Xiao et~al.(2023)Xiao, Tian, Chen, Han, and Lewis]{xiao2023efficient}
Xiao, G., Tian, Y., Chen, B., Han, S., and Lewis, M.
\newblock Efficient streaming language models with attention sinks.
\newblock \emph{arXiv preprint arXiv:2309.17453}, 2023.

\bibitem[Xiong et~al.(2023)Xiong, Liu, Molybog, Zhang, Bhargava, Hou, Martin, Rungta, Sankararaman, Oguz, et~al.]{xiong2023effective}
Xiong, W., Liu, J., Molybog, I., Zhang, H., Bhargava, P., Hou, R., Martin, L., Rungta, R., Sankararaman, K.~A., Oguz, B., et~al.
\newblock Effective long-context scaling of foundation models.
\newblock \emph{arXiv preprint arXiv:2309.16039}, 2023.

\bibitem[Xue et~al.(2020)Xue, Constant, Roberts, Kale, Al-Rfou, Siddhant, Barua, and Raffel]{xue2020mt5}
Xue, L., Constant, N., Roberts, A., Kale, M., Al-Rfou, R., Siddhant, A., Barua, A., and Raffel, C.
\newblock mt5: A massively multilingual pre-trained text-to-text transformer.
\newblock \emph{arXiv preprint arXiv:2010.11934}, 2020.

\bibitem[Yang et~al.(2023)Yang, Jin, Tang, Han, Feng, Jiang, Yin, and Hu]{yang2023harnessing}
Yang, J., Jin, H., Tang, R., Han, X., Feng, Q., Jiang, H., Yin, B., and Hu, X.
\newblock Harnessing the power of llms in practice: A survey on chatgpt and beyond.
\newblock \emph{arXiv preprint arXiv:2304.13712}, 2023.

\bibitem[{Yin Song and Chen Wu and Eden Duthie}(2023)]{MistralLite_2023}
{Yin Song and Chen Wu and Eden Duthie}.
\newblock {amazon/MistralLite}, 2023.
\newblock URL \url{https://huggingface.co/amazon/MistralLite}.

\bibitem[Zaheer et~al.(2020)Zaheer, Guruganesh, Dubey, Ainslie, Alberti, Ontanon, Pham, Ravula, Wang, Yang, et~al.]{zaheer2020big}
Zaheer, M., Guruganesh, G., Dubey, K.~A., Ainslie, J., Alberti, C., Ontanon, S., Pham, P., Ravula, A., Wang, Q., Yang, L., et~al.
\newblock Big bird: Transformers for longer sequences.
\newblock \emph{Advances in neural information processing systems}, 33:\penalty0 17283--17297, 2020.

\bibitem[Zellers et~al.(2019)Zellers, Holtzman, Bisk, Farhadi, and Choi]{zellers2019hellaswag}
Zellers, R., Holtzman, A., Bisk, Y., Farhadi, A., and Choi, Y.
\newblock Hellaswag: Can a machine really finish your sentence?
\newblock \emph{arXiv preprint arXiv:1905.07830}, 2019.

\bibitem[Zhang et~al.(2023)Zhang, Chao, Dhurandhar, Chen, Tajer, Xu, and Yan]{zhang2023neural}
Zhang, J., Chao, H., Dhurandhar, A., Chen, P.-Y., Tajer, A., Xu, Y., and Yan, P.
\newblock When neural networks fail to generalize? a model sensitivity perspective.
\newblock In \emph{Proceedings of the AAAI Conference on Artificial Intelligence}, volume~37, pp.\  11219--11227, 2023.

\bibitem[Zhang et~al.(2022)Zhang, Roller, Goyal, Artetxe, Chen, Chen, Dewan, Diab, Li, Lin, et~al.]{zhang2022opt}
Zhang, S., Roller, S., Goyal, N., Artetxe, M., Chen, M., Chen, S., Dewan, C., Diab, M., Li, X., Lin, X.~V., et~al.
\newblock Opt: Open pre-trained transformer language models.
\newblock \emph{arXiv preprint arXiv:2205.01068}, 2022.

\bibitem[Zhao et~al.(2023)Zhao, Zhou, Li, Tang, Wang, Hou, Min, Zhang, Zhang, Dong, et~al.]{zhao2023survey}
Zhao, W.~X., Zhou, K., Li, J., Tang, T., Wang, X., Hou, Y., Min, Y., Zhang, B., Zhang, J., Dong, Z., et~al.
\newblock A survey of large language models.
\newblock \emph{arXiv preprint arXiv:2303.18223}, 2023.

\bibitem[Zhu et~al.(2023)Zhu, Yang, Wang, Song, Wu, Wei, and Li]{zhu2023pose}
Zhu, D., Yang, N., Wang, L., Song, Y., Wu, W., Wei, F., and Li, S.
\newblock Pose: Efficient context window extension of llms via positional skip-wise training.
\newblock \emph{arXiv preprint arXiv:2309.10400}, 2023.

\end{thebibliography}
\bibliographystyle{icml2024}
\clearpage

\appendix
\onecolumn

\section{Pseudocode of SelfExtend}

 \begin{figure}[h]
 \centering
    \begin{minipage}{0.50\textwidth}
    \begin{algorithm}[H]
    \caption{\footnotesize PyTorch-style Pseudocode of SelfExtend}\label{alg:SelfExtend}
\begin{lstlisting}[language=python]
  q, k, v # queries, keys, and values
  seq_len, pos # input sequence length, position_idx
  g_size, w_size = G, w_n 
  
  # normal self-attention
  ngb_q = apply_pos_emcode(q, pos)
  ngb_k = apply_pos_emcode(k, pos)
  ngb_attn = matmul(ngb_q, ngb_k)
  ngb_attn = causal_mask(ngb_attn)
  
  # grouped self-attention
  g_pos = pos // g_size  # the floor operation
  shift = w_size - w_size // g_size
  s_g_pos = g_pos + shift
  g_q = apply_pos_emcode(q, s_g_pos)
  g_k = apply_pos_emcode(k, g_pos)
  g_attn = matmul(g_q, g_k)
  g_attn = causal_mask(g_attn)
  
  g_mask = tril(ones([seq_len-w_size, seq_len-w_size]))
  mask = ones([seq_len, seq_len])
  mask[w_size:, :-w_size] -= g_mask
  
  attn = where(mask, ngb_attn, g_attn) # merge by replacement
  
  attn_weights = softmax(attn)
  output = matmul(attn_weights, v)
\end{lstlisting}

    \end{algorithm}
    \end{minipage}
\end{figure}

\section{Perplexity as a Metric for Long Context Capabilities}
\label{apdx:ppl}
\begin{figure}[h]
\centering
\includegraphics[width=0.45\linewidth]{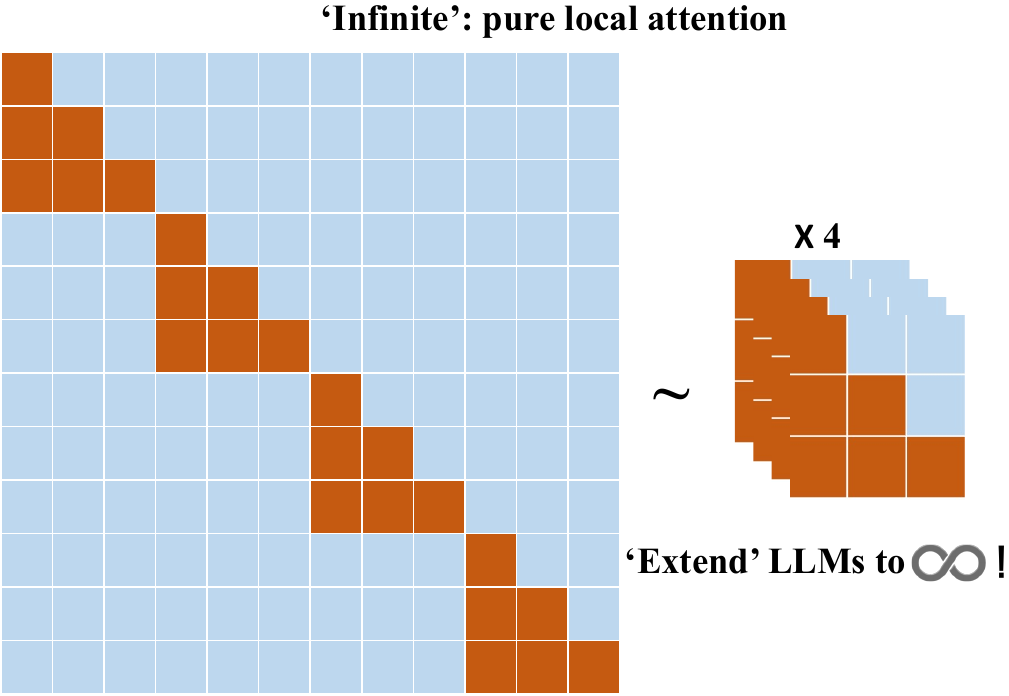}
\caption{'Infinite': a seemingly plausible method that extends an LLM's context window to 'infinite' length. It achieves this by dividing the entire self-attention area into multiple small, local self-attention areas. The size of the local window~(i.e., the spanning range) of a local self-attention area is the sole hyperparameter for "Infinite". For instance, with a local window set to 4 and a 16-token-long input, "Infinite" essentially processes the input as four sequences of 4 tokens each.}
\label{apdx:fig:local}
\end{figure}

PPL is not an effective metric for measuring the ability of LLMs to handle long contexts. In \cref{apdx:fig:local}, we introduce a seeming plausible context window extension method named 'Infinite'. When evaluated on PG19 using the same protocol, Llama-2-7b-chat with `Infinite' achieves PPL scores that are comparable to, or even lower than, those achieved by SelfExtend, as demonstrated in \cref{table:local}. However, `Infinite' essentially mimics the process of dividing a long sequence into short sub-sequences before processing them with LLMs, indicating that it does not genuinely address long context handling.

\begin{table}[h]
\centering
\caption{Perplexity on the PG19 dataset: For `Infinite', we set three different local window sizes: 1024, 2048, and 4096. We have also included the results from Table \ref{table:ppl} for comparison.}\label{table:local}
\resizebox{0.8\linewidth}{!}{
    \begin{tabular}{@{}lcccccccc@{}}
    \toprule
    Model & \multicolumn{6}{c}{Evaluation Context Window Size} \\ 
    \textbf{Name}   & \textbf{4096} &\textbf{6144} & \textbf{8192} & \textbf{10240} & \textbf{12288} &  \textbf{14336} & \textbf{16384} \\
    \bottomrule
    Llama-2-7b-chat & 9.181 &  $>10^3$ & $>10^3$ & $>10^3$ & $>10^3$ & $>10^3$ &$> 10^3$  \\
    SelfExtend-Llama-2-7b-chat  & 8.885 & 8.828 & 9.220 & 8.956 & 9.217 & 9.413 & 9.274 \\
    \hline
    1024-`Infinite'--Llama-2-7b-chat & 9.556 & 9.393 & 9.728 & 9.266 & 9.400 & 9.369 & 9.142 \\
    2048-`Infinite'--Llama-2-7b-chat & 9.288 & 9.045 & 9.478 & 8.993 & 9.128 & 9.105 & 8.872 \\
    4096-`Infinite'--Llama-2-7b-chat & 9.181 & 9.045 & 9.506 & 8.993 & 9.165 & 9.105 & 8.856 \\
    \bottomrule
    Mistral-7b-instruct-0.1 w/ SWA  & 9.295 & 9.197 & 9.532 & 9.242 & 9.198 & 9.278 & 9.294\\
    Mistral-7b-instruct-0.1 w/o SWA  & 9.295& 9.205 & 10.20 & 55.35 & $> 10^3$ & $> 10^3$ & $> 10^3$ \\
    SelfExtend-Mistral-7b-instruct-0.1  & 9.272 & 9.103 & 9.369 & 9.070 & 8.956 & 9.022 & 9.128  \\
    \bottomrule
    \end{tabular}
}
\end{table}
The discrepancy between Perplexity (PPL) and long context ability primarily stems from how PPL is calculated by averaging over numerous tokens. As long as the majority of tokens are modeled accurately, PPL will remain low. This is closely related to the influence of neighboring tokens. Information from neighboring tokens—such as those within the local attention window of 'Infinite'—can suffice for predicting most tokens, thus leading to a low PPL. However, a few critical tokens, which are crucial for understanding long contexts and answering questions, may not be predicted accurately.

Additionally, unlike the pre-training process where the cross-entropy loss corresponds directly to perplexity, measuring PPL during inference is static. It resembles a specific point on the loss curve observed during pre-training. While a decreasing trend in loss during pre-training indicates good performance, a single point on the training loss curve cannot determine the performance.

In summary, while low PPL is essential for a good model, lower PPL does not necessarily equate to better performance in understanding long contexts.

\section{SelfExtend with Varying Group Size and Neighbor Window}
\label{apdx:benchmark}
\begin{figure}[h!]
\centering
\vspace{-10pt}
\includegraphics[width=\linewidth]{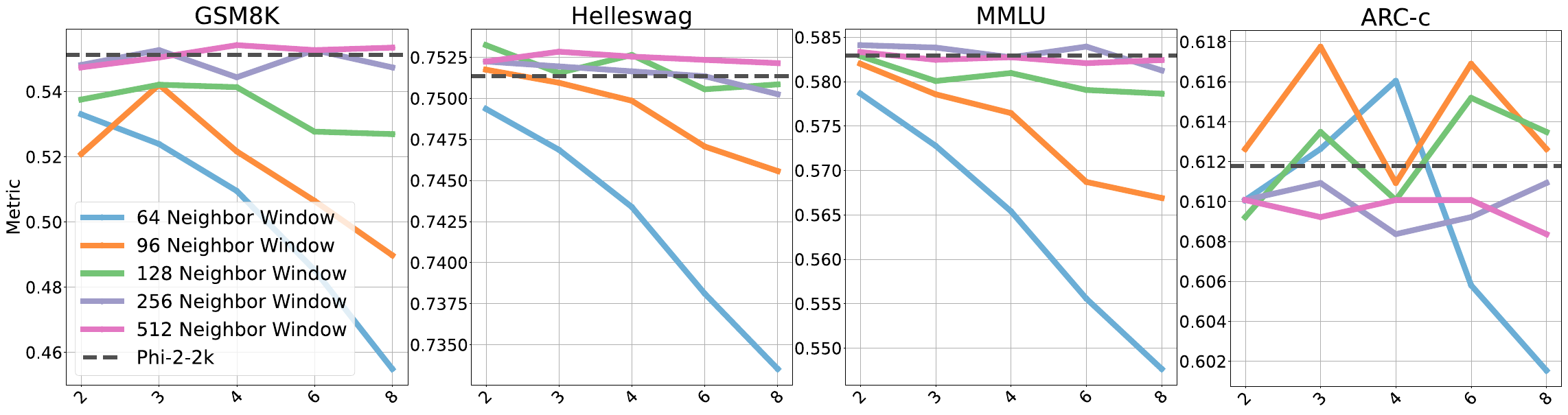}
\vspace{-10pt}
\caption{Phi-2 with SelfExtend on GSM8K, Helleswag, MMLU and ARC-c, compared to the vanilla Phi-2~(Phi-2-2k). The x-axis shows the group size and the y-axis indicates performance as measured by the corresponding metrics.}
\label{apdx:fig:short_bench}
\end{figure}

To comprehensively understand SelfExtend's influence on LLMs, unlike previous experiments which used long context settings, we evaluate with smaller neighbor window sizes on four standard benchmark tasks: ARC-c, GSM8k, Hellaswag and MMLU. We use Phi-2 as the extended LLM. The results are shown in \cref{apdx:fig:short_bench}. We didn't include TruthfulQA because its average length is less than 300 words, while the four datasets we used have an average length greater than 700 words. In general, SelfExtend has a minor influence on Phi-2 as long as the neighbor window size is over 128. In many cases, SelfExtend even performs slightly better than vanilla Phi-2. When the neighbor window is too small (e.g. 64 tokens), if the group size is large, as expected, the positional information loss is too high and Phi-2's performance degrades. Also, on difficult tasks such as MMLU and Helleswag, we observe a monotonic decrease in performance with increasing group size for all neighbor windows. In summary, even when applying SelfExtend to short context tasks, as long as the hyperparameters are not extreme, SelfExtend does not harm the model.

\section{Detailed Experimental Setting}\label{sec:app:setting}

In this appendix, we present the details of the experiments in our paper.

\subsection{Experimental Setting on Language Modeling Tasks}\label{sec:app:setting:llt}

This is not the standard setting for PPL testing on PG-19. We use the first sentence of each book in PG19's test set~(100 books) to test the language modeling ability. The results cannot be directly compared to the PPL reported by other papers. We chose this setting because our computation resources are very limited. This setting saves a lot and it can still show the behavior of LLMs w.r.t. PPL. All PPL results were calculated using the sliding window method~\cite{press2021train} with $S = 256$. We evaluated how the PPL changes as the input length increases. In \cref{table:ppl}, SelfExtend extends the original Llama-2's context window length from $4096$~(4k) to over $16384$~(16k) with group size~$G_s$ set as $8$ and neighbor window~$w_n$ set as $1024$~(1k). For Mistral model, without SWA, the context window is $8192$~(8k) and it is also extended by SelfExtend with the same setting to larger than 16k. With SWA, Mistral can digest an infinite length of sequences and its default sliding window is $4096$.

 \subsection{Experimental Setting on Passkey Retrieval Task}\label{apdx:passkey_example}

Compared to other synthetic tasks, such as ''Needle in a Haystack''~\cite{gkamradt2023needle}, the model's performance on this is not sensitive to the prompt~\cite{needle_claudes}. This may come from the fact that the sentence carrying the passkey is very different from those repeated random texts surrounding it. Empirically, within the effective context window, almost all LLMs, including those without any instruction tuning or alignment, can locate the sentence carrying the passkey. Although this task is easy and far from real-world scenarios, it tests two fundamental capabilities of LLMs: 1. The model should be able to recognize and locate the useful information across all positions of the input sequence~(the most fundamental understanding capability); 2. The model should be able to use the perceived information to finish tasks~(the most fundamental generation capability).

An example of passkey is as the following:

\begin{figure}[h!]
\centering
\fcolorbox{black}{gray!10}{\parbox{.9\linewidth}
    {
        \hspace{-1cm}\textbf{Example:} \\
        \textbf{Prompt:} There is an important info hidden inside a lot of irrelevant text. Find it and memorize it. I will quiz you about the important information there …… back again. The grass is green. The sky is blue. The sun is yellow. Here we go. There and back again.The grass is green. The sky is blue. The sun is yellow. Here we go. There and back again.The grass is green. The sky is blue. The sun is yellow. Here we go. There and back again.\underline{The pass key is 60151. Remember it. 60151 is the pass key.} The grass is green. The sky is blue. The sun is yellow. Here we go. There and back again.The grass is green. The sky …… What is the passkey?\\
        \textbf{Ground Truth:} 60151 \\
        \hangindent=1cm \hangafter=0
    }
}
\caption{An example of the passkey retrieval task.}
\label{apdx:fig:passkey}
\end{figure}

\subsection{Experimental Setting on Varying-Length Passkey Retrieval Task}\label{sec:app:setting:lprt}

In this experiment, we use the following models: Llama2-7b-chat with SelfExtend, LongLora-7b-16k\footnote{We use its fully fine-tuned variant, as we cannot use the LongAlpaca version to get reasonable performance for this specific task. For more details about the model: https://huggingface.co/Yukang/Llama-2-7b-longlora-16k},vicuna-1.5-7b-16k, Together AI's Llama-2-7b-32k\footnote{Both vicuna-1.5-7b-16k and Together AI's Llama-2-7b-32k were fine-tuned using position interpolation}, and Yarn-Llama-2-7b-64k.

\section{Detail of LLMs}\label{sec:app:models}

Here, we list the links to the details of the LLMs utilized in our experiments.

\begin{table}[H]
\centering
\caption{LLMs used in the experiments}\label{table:ai_models}
\begin{tabularx}{\textwidth}{lX}
\hline
Model Name & URL \\
\hline
Llama-2-7b-chat-hf~\cite{touvron2023llama} & \url{https://huggingface.co/meta-llama/Llama-2-7b-chat-hf} \\
Mistral-7B-Instruct-v0.1~\cite{jiang2023mistral} & \url{https://huggingface.co/mistralai/Mistral-7B-Instruct-v0.1} \\
Phi-2~\cite{javaheripi2023phi} & \url{https://huggingface.co/microsoft/phi-2} \\
SOLAR-10.7B-Instruct-v1.0~\cite{kim2023solar} & \url{https://huggingface.co/upstage/SOLAR-10.7B-Instruct-v1.0} \\
LongChat-7b-v1.5-32k & \url{https://huggingface.co/lmsys/longchat-7b-v1.5-32k} \\
togethercomputer/LLaMA-2-7B-32K & \url{https://huggingface.co/togethercomputer/LLaMA-2-7B-32K} \\
CLEX-7B-16K~\cite{chen2023clex} & \url{https://huggingface.co/DAMO-NLP-SG/CLEX-7B-16K} \\
CodeLlama-7b-hf~\cite{roziere2023code} & \url{https://huggingface.co/codellama/CodeLlama-7b-hf} \\
vicuna-7b-v1.5-16k & \url{https://huggingface.co/lmsys/vicuna-7b-v1.5-16k} \\
MistralLite~\cite{amazon2023mistrallite} & \url{https://huggingface.co/amazon/MistralLite} \\
\hline
\end{tabularx}

\end{table}

\section{SelfExtend with Other Positional Encodings}
In this section, we test SelfExtend on LLMs using non-RoPE positional encodings. We implemented SelfExtend for MPT-7b-chat~\cite{MosaicML2023Introducing}, which uses Alibi~\cite{press2021train} as its positional embedding method. We conducted experiments on the PG19 dataset. The results are shown in \cref{tab:alibi}. The results show that SelfExtend is able to work with non-RoPE positional encodings, which are pretty similar to models with RoPE positional encoding,
\begin{table}[h]
    \centering
    \caption{Perplexity of MPT-7b-chat on PG19 with different sequence length. The vanilla MPT-7b-chat has a context window of 2k tokens. For SelfExtend, We set the neighbor window as 512 and set the group size as 6.}\label{tab:alibi}
    \begin{tabular}{l|c|c|c|c|c|c|c}
         \toprule
         MPT-7b-chat & 1024 & 2048 & 3172 & 4096 & 5120 & 6144 & 8192 \\
         \midrule
         Vanilla~(2k) & 8.8 & 9.6 & 12.0 & 26.3 & 52.0 & 115.5 & 196.0 \\
         SelfExtend & 8.9& 9.9 &10.6& 10.8 &11.1 &11.3 & 11.6 \\
         \bottomrule
    \end{tabular}

\end{table}

\section{Hyperparameyer Selection for SelfExtend}
\label{apdx:hypara}

\begin{figure}[h]
    \centering
    \includegraphics[width=\linewidth]{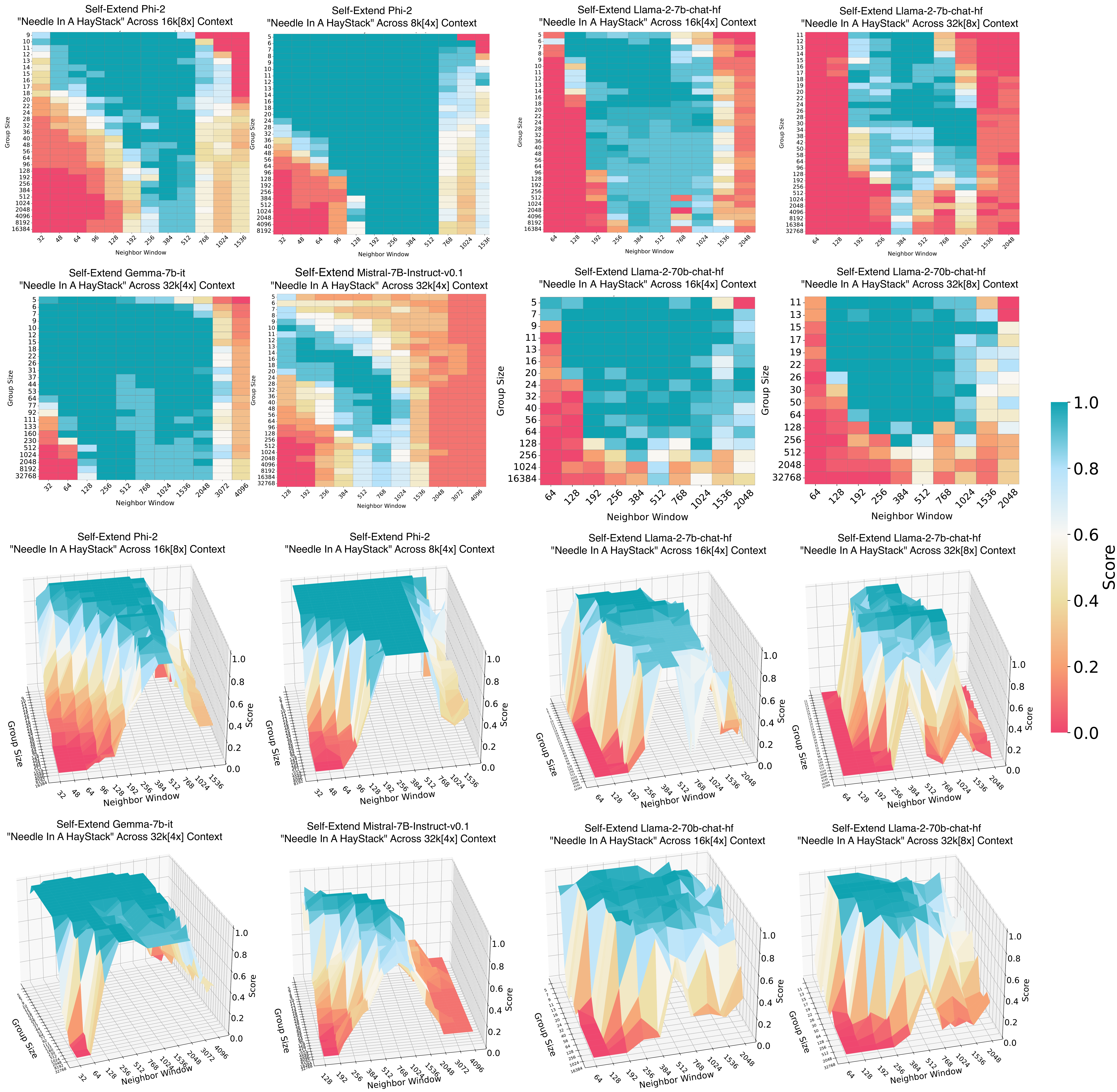}
    \caption{Impacts of group size and neighbor window size}
    \label{fig:hypara}
\end{figure}

We conduct experiments on ``Needle in a Haystack''~\cite{gkamradt2023needle}, to investigate the impacts of group size and neighbor window size. The results are shown in \cref{fig:hypara}. 

The experimental results indicate that SelfExtend is not overly sensitive to hyperparameter selection. Predefined, heuristic values for group size and neighbor window size are often sufficient to achieve satisfactory performance, as long as group size and neighbor window are not too large or too small. We conclude those results as an empirical rule. Denoting the pretraining context window as $L$, the target extension length as $N$, the neighbor window as $W$, and the group size as $G$, the empirical rule for selecting hyperparameters is to ensure that the following inequality holds: 
\begin{equation}
    \frac{1}{2} \times L > W + \frac{N-W}{G}
\end{equation}
We believe this empirical rule is due the fact that: large relative positions are not well trained. Empirically, only a portion($\sim\frac{1}{2}$) of positions are well-trained and SelfExtend should only leverage these well-trained relative positions for the extension. This finding explains: excessively small group sizes can degrade performance, as they provide precise position information but require SelfExtend to utilize less well-trained relative positions for extension; excessively large neighbor window sizes can also degrade performance, as they provide more neighbor information but necessitate the use of less well-trained relative positions for extension.

However, the current observation may not be applicable to all models. For example, we've found that Llama3 series should use a much smaller neighbor window~($\sim 100$). We may dive deeper to investigate the interaction among those hyperparameters and models. Besides following the empirical rule. One could use a simple and easy-to-run representative task to find proper hyperparameters.

\end{document}